\documentclass[11pt, a4paper, logo, twocolumn, copyright, nonumbering]{googledeepmind}

\usepackage[utf8]{inputenc}
\usepackage{amssymb}
\usepackage{framed,enumitem} 
\usepackage{amsmath}
\usepackage{amsxtra}
\usepackage{mathtools}
\usepackage{graphicx}
\usepackage{hyperref}
\usepackage{bookmark}
\usepackage{booktabs}
\usepackage{nicematrix}
\usepackage{listings}
\usepackage{xcolor}
\usepackage{float}
\usepackage{cuted} 
\usepackage[super,comma,sort&compress]{natbib} 
\usepackage[all]{nowidow}
\definecolor{codegreen}{rgb}{0,0.6,0}
\definecolor{codegray}{rgb}{0.5,0.5,0.5}
\definecolor{codepurple}{rgb}{0.58,0,0.82}
\definecolor{backcolour}{rgb}{0.95,0.95,0.92}

\lstdefinestyle{mystyle}{
    backgroundcolor=\color{backcolour},   
    commentstyle=\color{codegreen},
    keywordstyle=\color{magenta},
    numberstyle=\tiny\color{codegray},
    stringstyle=\color{codepurple},
    basicstyle=\footnotesize,
    breakatwhitespace=false,         
    breaklines=true,                 
    captionpos=b,                    
    keepspaces=true,                 
    numbers=left,                    
    numbersep=5pt,                  
    showspaces=false,                
    showstringspaces=false,
    showtabs=false,                  
    tabsize=2
}

\DeclareMathAlphabet{\doublestruck}{U}{BOONDOX-ds}{m}{n}
\newcommand{\irreps}[1]{\doublestruck{#1}}
\newcommand{\irrep}[2]{\doublestruck{#1}_{#2}}
\newcommand{\couples}[1]{\otimes^{({#1})}}
\newcommand{\couple}[2]{\otimes^{({#1}_{#2})}}

\definecolor{hartmut}{rgb}{0,.6,.6}
\definecolor{oliver}{rgb}{.6,0,0}

\renewcommand{\vv}[1]{\boldsymbol{#1}}
\newcommand{\mat}[1]{\boldsymbol{#1}}
\newcommand\T{{\mathpalette\raiseT\intercal}}
\newcommand\raiseT[2]{\raisebox{0.25ex}{$#1#2$}}
\newcommand{\arrow}{\rightarrow}%
\newcommand{\BR}{{\mathbb R}}%
\newcommand{\BS}{{\mathbb S}}%
\newcommand{\Ethree}{{\mathrm{E}(3)}}
\newcommand{\SEthree}{{\mathrm{SE}(3)}}
\newcommand{\Othree}{{\mathrm{O}(3)}}
\newcommand{\SOthree}{{\mathrm{SO}(3)}}
\newcommand{\SOtwo}{{\mathrm{SO}(2)}}

\providecommand\phantomsection{}

\hypersetup{bookmarksdepth=4}

\title{\texttt{E3x}: $\Ethree$-Equivariant Deep Learning Made Easy}

\correspondingauthor{oliverunke@google.com}

\reportnumber{} 

\renewcommand{\today}{}

\author[1]{Oliver T. Unke}
\author[1]{Hartmut Maennel}

\affil[1]{Google DeepMind}

\begin{abstract}
This work introduces \texttt{E3x}, a software package for building neural networks that are equivariant with respect to the Euclidean group $\Ethree$, consisting of translations, rotations, and reflections of three--dimensional space. Compared to ordinary neural networks, $\Ethree$-equivariant models promise benefits whenever input and/or output data are quantities associated with three--dimensional objects. This is because the numeric values of such quantities (e.g.\ positions) typically depend on the chosen coordinate system. Under transformations of the reference frame, the values change predictably, but the underlying rules can be difficult to learn for ordinary machine learning models. With built-in $\Ethree$-equivariance, neural networks are guaranteed to satisfy the relevant transformation rules \emph{exactly}, resulting in superior data efficiency and accuracy.\\ The code for \texttt{E3x} is available from \url{https://github.com/google-research/e3x}, detailed documentation and usage examples can be found on \url{https://e3x.readthedocs.io}. 
\end{abstract}

\begin{document}

\maketitle

\phantomsection
\section{Introduction}
\label{sec:introduction}

When we want to describe some geometric object in space (e.g.\ a collection of points or a distribution of masses) with numbers, we need to choose a reference point as origin and an orthonormal basis of vectors in $x$, $y$, and $z$ direction. This defines a coordinate system, which identifies points in physical space with mathematical points in $\BR^3$. However, this coordinate system usually is arbitrarily chosen, we could as well use any rotated (or translated) version of it. When we process geometric objects with a machine learning model to extract a classification or number (for regression tasks), the result should usually not depend on our choice of coordinate system, so it should be \emph{invariant} under translations, rotations, and (depending on the task), maybe also reflections.

We may also want to extract a vector, e.g.\ if the geometric objects are cameras, we may want to assign the direction in which this camera points, or if it is a physical object, we may want to predict the forces acting on it. In those cases, the numerical representation of the vector changes if we e.g.\ rotate the coordinate system, but ``in the same way'' as the coordinates of the geometric objects. Similarly, we may want to extract other properties, like the direction of a symmetry axis, or an approximating ellipsoid, which also ``change accordingly'' when the coordinate system is changed. Such properties are not invariant, but \emph{equivariant}. 

If we want to use a multi-layer neural network to predict such invariant or equivariant properties, we can make sure that the end result is invariant/equivariant by requiring that each layer in the network works with ``equivariant features'' and all operations respect this equivariance. Even if we are only interested in end results that should be invariant numbers, it is often useful to use equivariant features as intermediates. For example, consider equivariant ``vector features''. To extract an invariant output from such features, we can compute their scalar (dot) product. Since the dot product of two vectors contains information about the angle between them, this allows to efficiently resolve ``angular information'' in a manner that would not be possible without intermediate equivariant features.

In recent years, equivariant machine learning models\cite{cohen2018spherical,thomas2018tensor,weiler20183d,cohen2019general,fuchs2020se} have been successfully applied in many different fields, e.g.\ in computer vision,\cite{cohen2016group,marcos2017rotation} mesh reconstruction,\cite{chatzipantazis2022se} and quantum chemistry.\cite{anderson2019cormorant,unke2021se,schutt2021equivariant,unke2021spookynet,batzner20223} However, implementing equivariant operations is non-trivial and can be difficult to reconcile with existing neural network building blocks. In practice, this often means that intuition and skills acquired by designing ordinary neural networks do not carry over to building equivariant models, and it can be difficult to adapt an existing model architecture to an equivariant version. In this work, we introduce \texttt{E3x}, a library for building $\Ethree$--equivariant neural networks built on Flax,\cite{flax2020github} with the aim to make the construction of equivariant models intuitive and simple. This is achieved by generalizing features and neural network building blocks to be equivariant in a way that allows recovering ordinary features and neural network behaviour as a limiting case.

The ``ingredients'' of the above-mentioned equivariant features are discussed in detail in the \nameref{sec:mathematical_background}. This section is meant as a self-contained introduction to the relevant mathematical theory, a learning resource, and a quick reference for formal definitions of technical terms, such as equivariance. Readers that are already familiar with the topic, or primarily interested in the practical implementation, may want to skip directly to \nameref{sec:how_e3x_works} for an overview of how equivariant features and operations in \texttt{E3x} are designed concretely.

\phantomsection
\section{Mathematical background}
\label{sec:mathematical_background}

In the following, we give formal definitions for the most important mathematical concepts and an overview of the theory underlying \texttt{E3x}. Beyond abstract definitions, whenever possible, we aim to give readers an intuitive understanding of the relevant concepts with concrete examples. We also motivate why the construction of equivariant features leads to these concepts and why all information that can be contained in equivariant features can be built up from functions that are the product of a radial function and so-called spherical harmonics.

\phantomsection
\subsection[Groups]{Groups and group actions}
\label{sec:groups}

\begin{table}
\centering
\begin{tabular}{c l}
\toprule
\textbf{group} & \textbf{transformations} \\
\midrule
$\Ethree$ & translations, rotations, reflections \\
$\SEthree$ & translations, rotations \\
$\Othree$ & rotations, reflections \\
$\SOthree$ & rotations \\
\bottomrule
\end{tabular}
\caption{Overview of groups and transformations relevant for three--dimensional objects.}
\label{tab:groups_relevant_in_3d}
\end{table}

\phantomsection
\paragraph{Definition}
\label{sec:groups_definition}
A group $G$ is a non-empty set\footnote{A \emph{set} is the mathematical model for a collection of things, which are also called \emph{elements} of the set.} equipped with a binary operation $\odot$ that combines any two elements $a,b$ of $G$ in such a way that
\vspace{-\topsep}
\begin{enumerate}
    \item $G$ is closed under the operation:\\
     $a \odot b$ is also an element of $G$,\vspace{0.25\baselineskip}
    \item the operation is associative:\\
    $a \odot (b \odot c) = (a \odot b) \odot c$,\vspace{0.25\baselineskip}
    \item an identity element $e$ exists:\\
    $a \odot e = e \odot a = a$,\vspace{0.25\baselineskip}
    \item and every element $a$ has an inverse $a^{-1}$:\\
    $a \odot a^{-1} = e$
\end{enumerate}
\vspace{-\topsep}
(see e.g.\ Chapter~1 of Ref.~\citenum{kosmann2009}). For us, $G$ will usually be one of the concrete groups $\mathrm{E(3)}$, $\mathrm{SE(3)}$, $\Othree$, $\SOthree$ shown in Table~\ref{tab:groups_relevant_in_3d}. In particular, these groups are topological groups, i.e.\ there is also a notion of continuous functions from $G$ to the real numbers $\BR$ or to vectors or matrices (or higher-order tensors), and with this the operation $\odot$ and inverse are continuous functions.

For example, rotations in the two--dimensional plane and in three--dimensional space are the elements of  groups called $\SOtwo$ and $\SOthree$, respectively. Rotations intuitively satisfy the four required properties of a group listed above: When two rotations $a$ and $b$ are applied consecutively to an object, the combined transformation $a \odot b$ is still a rotation (closure). Similarly, when applying a chain of multiple rotations to an object, any two consecutive rotations can first be combined without changing the overall result (associativity). Further, there is a ``rotation by zero degrees'' that does not change the orientation of objects (existence of identity element) and the effect of any rotation can be reverted by ``rotating back'' (existence of inverse element).

\phantomsection
\paragraph{Subgroups}
\label{sec:subgroups}
A subset $H\subseteq G$ is called a \emph{subgroup} of $G$ if it is closed under the group operation and under taking inverses (thus it also contains the identity element), so it automatically becomes a group in its own right. 

For example, the planar rotations around the $z$--axis can be identified with $\SOtwo$, which is a subgroup of the three-dimensional rotations in $\SOthree$. The rotations $\SOthree$ are themselves a subgroup of $\Othree$, which also includes rotoreflections (see Table~\ref{tab:groups_relevant_in_3d}).

\phantomsection
\paragraph[Group actions]{Action of a group on a set}
\label{sec:group_actions}
A group $G$ acts on a set $X$, if for every $g\in G, x\in X$ we have defined ``how to apply $g$ to $x$'', i.e.\ we have defined a $g x$, such that for all $g,h\in G$:
\begin{equation*}
   h(gx) = (h\odot g)x\,.
\end{equation*}
In other words, for every $g\in G$ there is a map \mbox{$\rho(g):X\arrow X$} such that \mbox{$\rho(h) \rho(g) = \rho(h\odot g)$}.

For example, the group $G=\SOthree$ can operate on $X=\BR^3$. The map \mbox{$\rho(g):\BR^3\arrow\BR^3$} would assign any vector $\vv r\in\BR^3$ to the corresponding rotated vector.

\phantomsection
\paragraph{Equivariance}
\label{sec:equivariance}
Let a group $G$ act on the sets $X$ and $Y$. A map \mbox{$f:X\arrow Y$} is called \emph{equivariant} if \mbox{$f(gx) = g(f(x))$} for all $g\in G$, $x\in X$, $f(x) \in Y$.

In other words, it does not matter whether a group element acts on the input or output of an equivariant function, the result will be the same. For example, consider the group $G=\SOtwo$. The functions
\begin{equation*}
n(\vv u) \coloneqq  \lambda \cdot \begin{bmatrix} x \\ y \end{bmatrix} \quad \text{and} \quad 
t(\vv u) \coloneqq \lambda \cdot \begin{bmatrix} -y \\ x \end{bmatrix}
\end{equation*}
(with some fixed $\lambda \neq 0 \in \BR$) are examples of equivariant functions $\BS^1 \arrow \BR^2$. Here, $n$ maps points $\vv u = [x \ y]^\T$ on the unit circle $\BS^1$ to vectors $\vv r \in \BR^2$ of length $\lambda$ pointing in radial (normal) direction, whereas $t$ maps $\vv u$ to vectors pointing in tangential direction (see illustrations below).
\begin{minipage}{.5\columnwidth}
\centering
\begin{equation*}
n(\vv u)
\end{equation*}
\includegraphics[width=\textwidth]{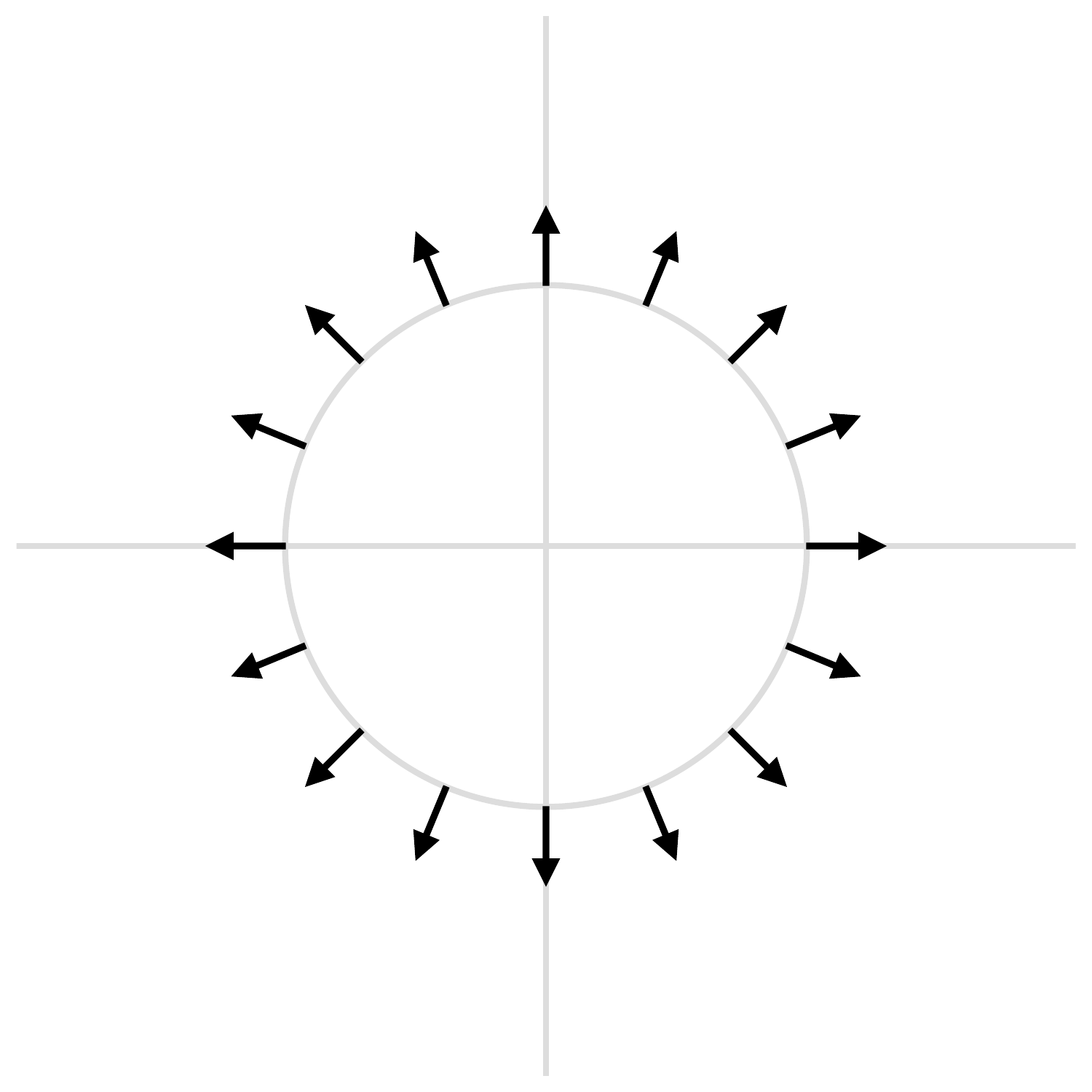}
\end{minipage}%
\begin{minipage}{.5\columnwidth}
\centering
\begin{equation*}
t(\vv u)
\end{equation*}
\includegraphics[width=\textwidth]{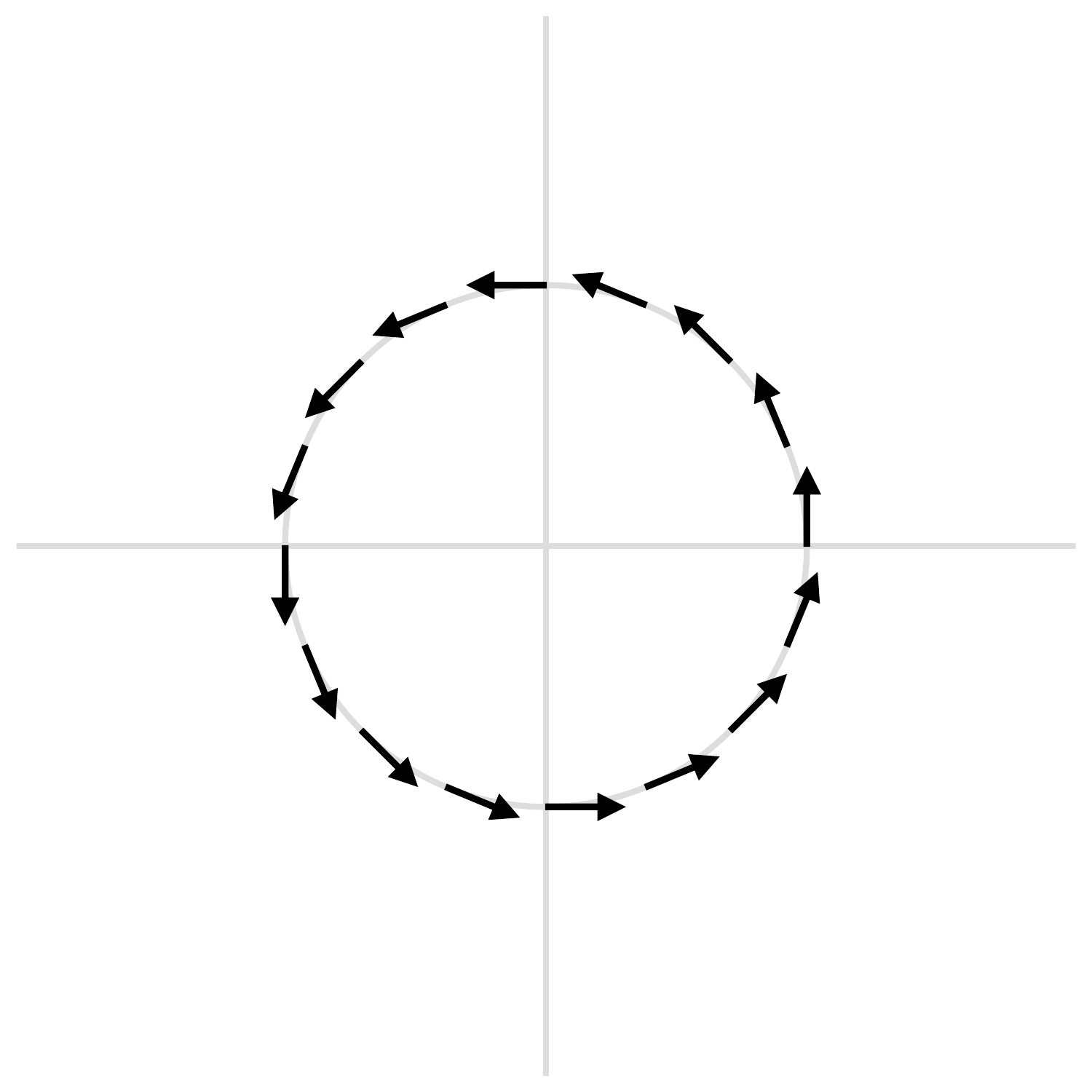}
\end{minipage}\\

We can easily extend $n$ and $t$ to equivariant functions $\BR^2 \arrow \BR^2$, for example by using the unit vector $\vv{\hat{r}} = \vv r/r$ as input to $n$ or $t$ and scaling the result by the radius $r=\lVert \vv r \rVert$ (see illustrations below).
\begin{minipage}{.5\columnwidth}
\centering
\begin{equation*}
\tilde n(\vv r) \coloneqq r \cdot n(\vv{\hat{r}})
\end{equation*}
\includegraphics[width=\textwidth]{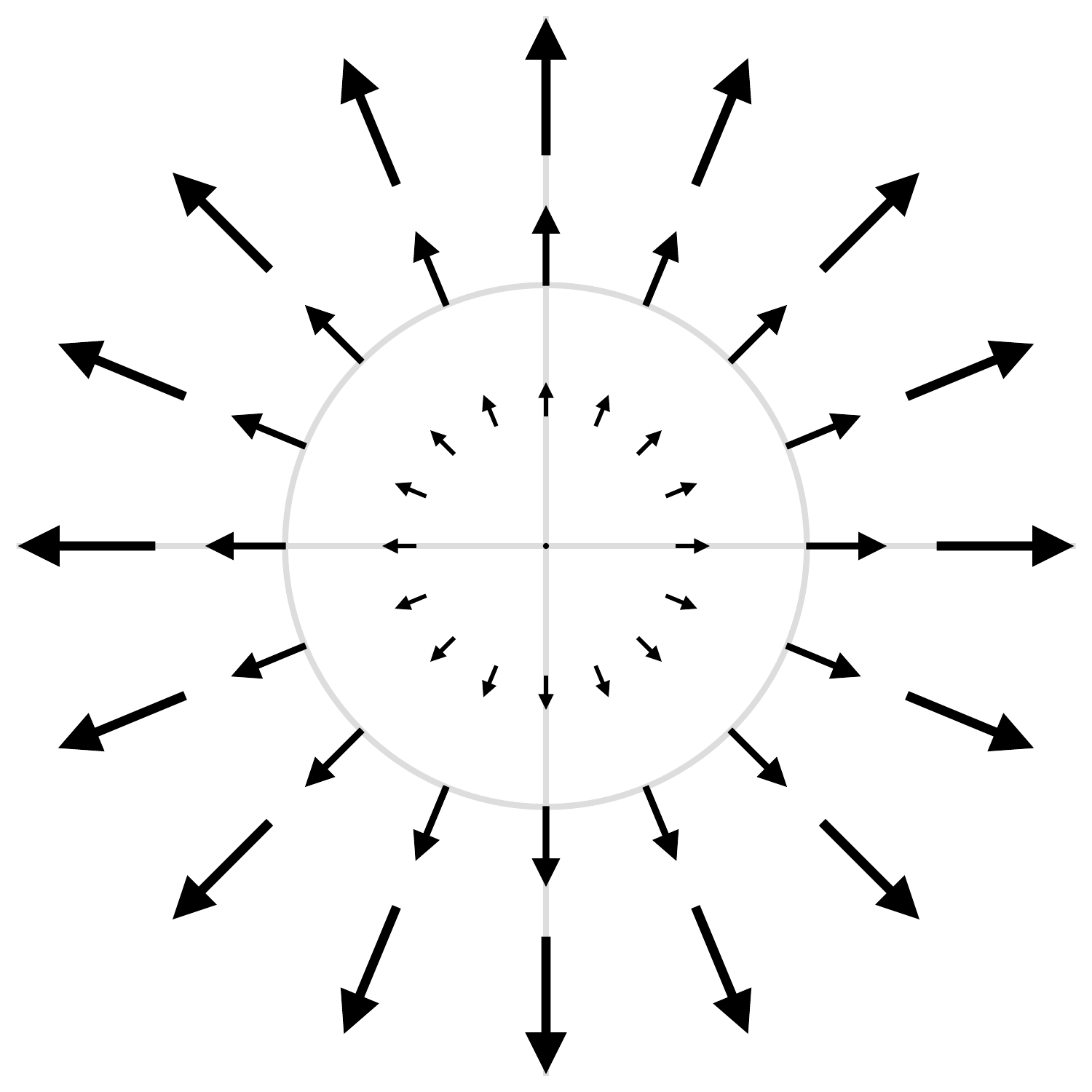}
\end{minipage}%
\begin{minipage}{.5\columnwidth}
\centering
\begin{equation*}
\tilde t(\vv r) \coloneqq r \cdot t(\vv{\hat{r}} )
\end{equation*}
\includegraphics[width=\textwidth]{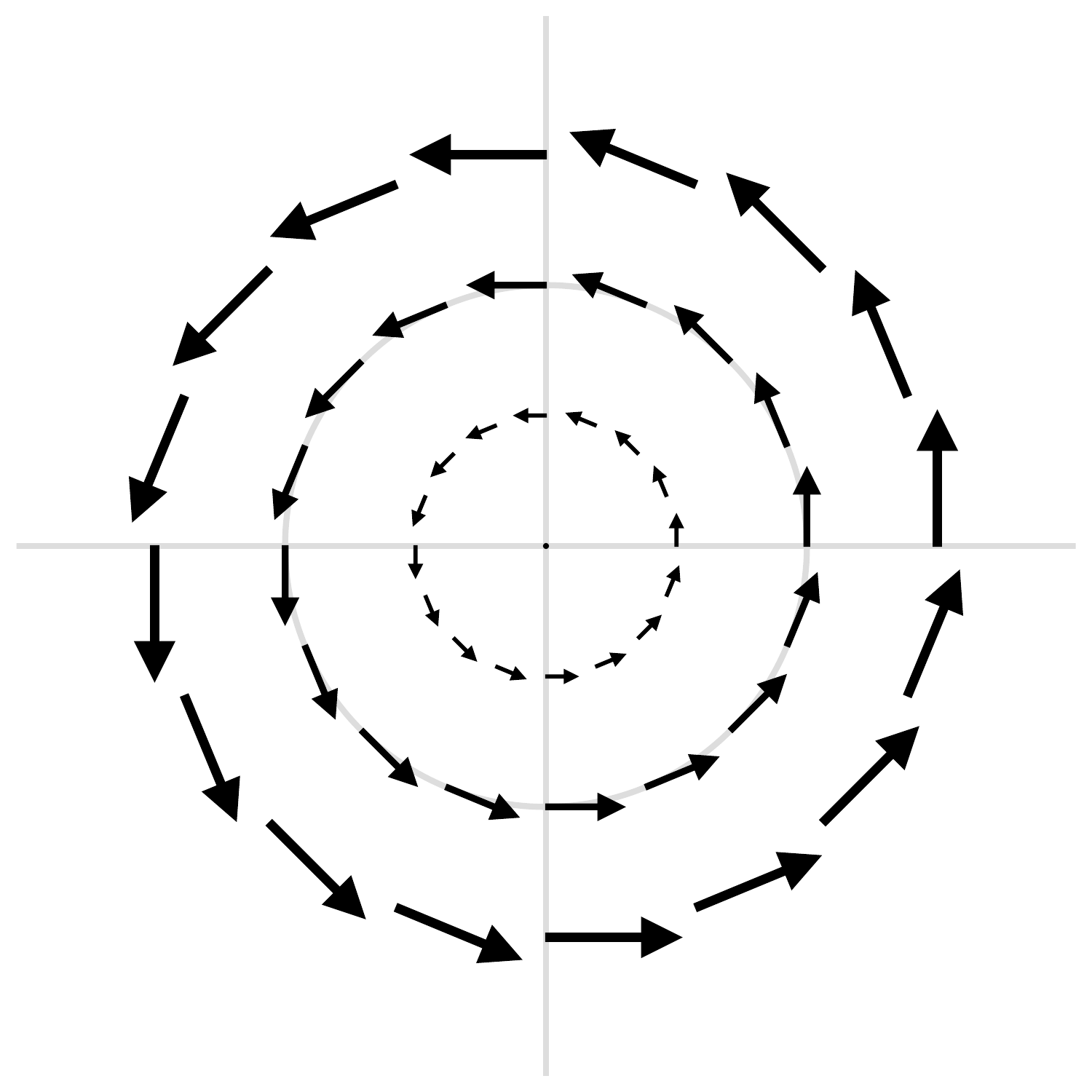}
\end{minipage}\\

More generally, we can use any continuous functions of the radius $a,b: \BR_{\geq 0} \arrow \BR$ satisfying $a(0)=b(0)=0$\footnote{The requirement $a(0)=b(0)=0$ is necessary for our formula, because the unit vector $\vv{\hat{r}}$ is undefined for $\vv r= \vv 0$. But also, \emph{any} equivariant continuous function $f:\BR^2\arrow \BR^2$ \emph{must} map the zero vector to itself, since for any rotation $g\in \SOtwo$ we have $g \vv 0 = \vv 0$, so we must have $g (f(\vv 0)) = f(\vv 0)$ and this is only possible for $f(\vv 0)=\vv 0$.} to define an equivariant continuous function $f:\BR^2\arrow \BR^2$ as
\begin{equation}
f(\vv r) \coloneqq  
a(r) \cdot n(\vv{\hat{r}}) + 
b(r) \cdot  t(\vv{\hat{r}})\,.
\label{eq:radial}
\end{equation}

In fact, any continuous $\SOtwo$--equivariant function $f:\BR^2\arrow \BR^2$ can be written in this form. For example, we might have an equivariant function that, close to the origin, maps points to short vectors pointing in radial direction and smoothly transitions to longer vectors rotated clockwise (in relation to the radial direction) further from the origin. We can choose $a,b$ in \eqref{eq:radial} such that the desired behaviour is satisfied for points in a specific direction, e.g.\ for those on the right half of the $x$--axis. Then, $f(\vv r)$ also \emph{must have} the desired behavior for all other points by equivariance (see illustration below).

\begin{minipage}{.5\columnwidth}
\centering
\includegraphics[width=\textwidth]{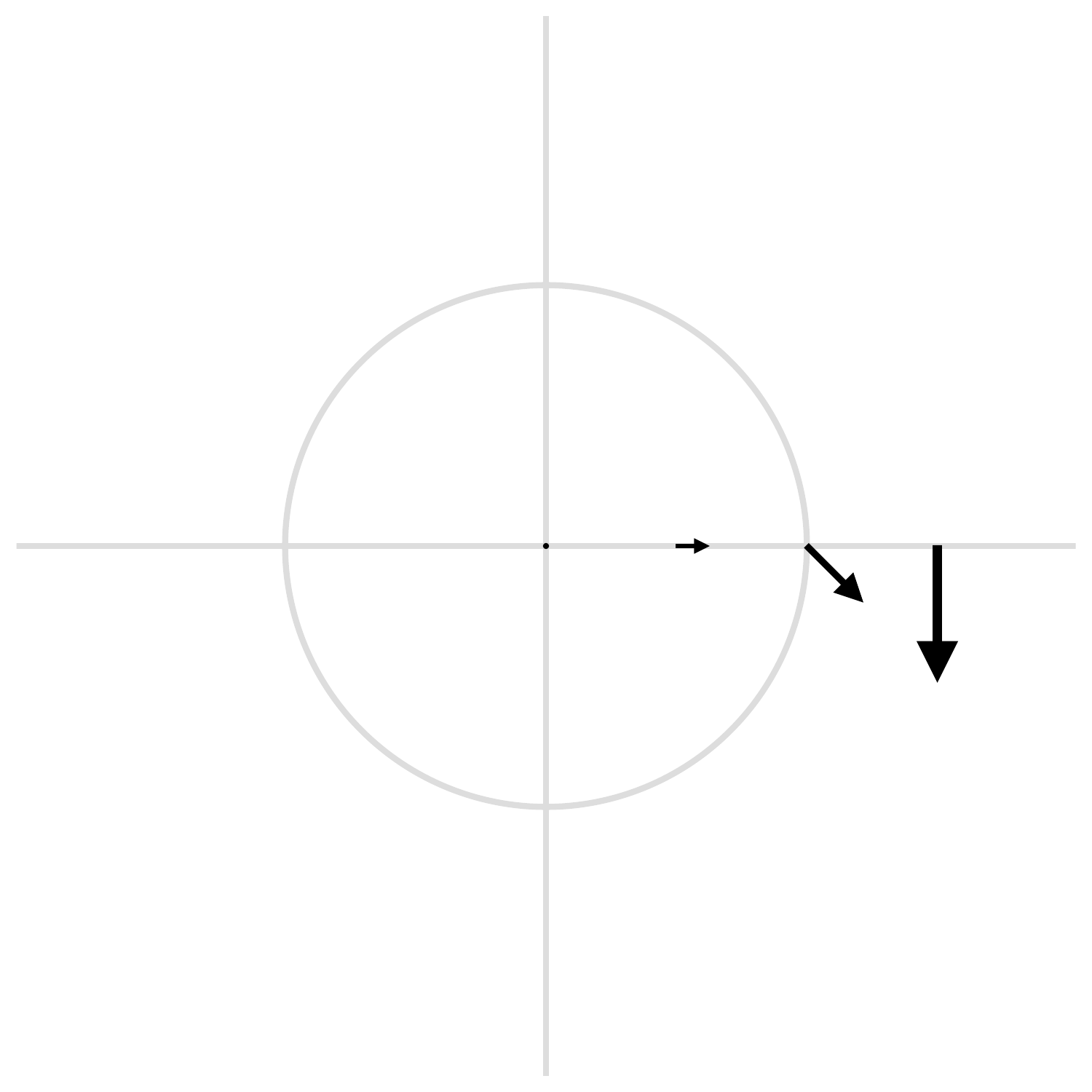}
\end{minipage}%
\begin{minipage}{.5\columnwidth}
\centering
\includegraphics[width=\textwidth]{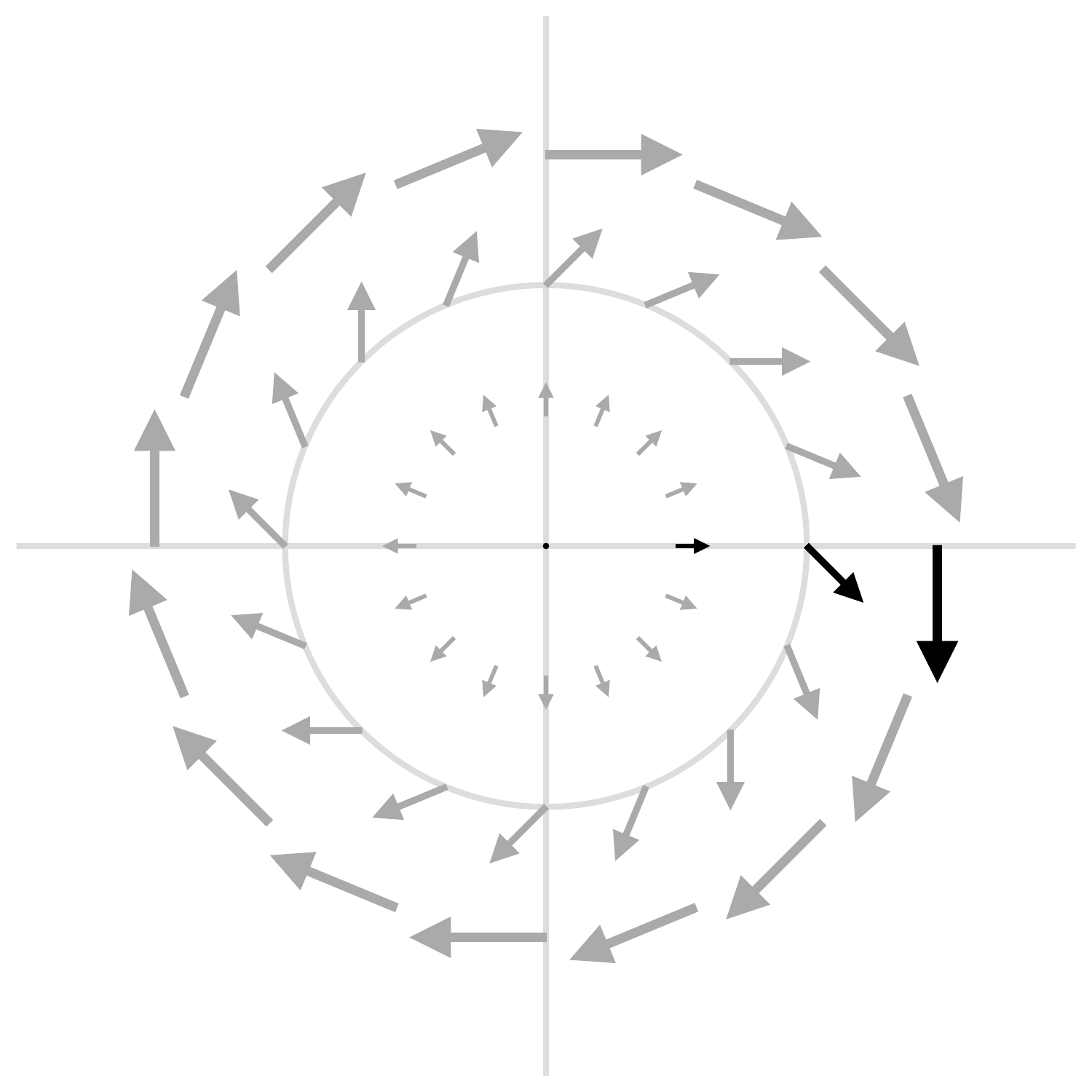}
\end{minipage}\\

Later, we use a similar construction to describe $\SOthree$--equivariant features on $\BR^3$ by equivariant functions on the three--dimensional unit sphere $\BS^{2}\subset\BR^3$ and radial functions $\BR_{\geq 0}\arrow \BR$.

Note that in \eqref{eq:radial}, we multiplied $n(\vv{\hat{r}})$ and $t(\vv{\hat{r}})$ with scalars $a(r)$ and $b(r)$ and added them together. For such expressions to make sense, the equivariant feature values $f(\vv r)$ must have the structure of a \emph{vector space}, which is defined in the next section. To formally show that \eqref{eq:radial} defines an equivariant function, we would use that group elements $g\in G$ act on this vector space by linear maps. Actions of a group on a vector space by linear maps are called \emph{representations}, which we study in detail after the next section.

\phantomsection
\subsection[Vector spaces]{Vector spaces}
\label{sec:vector_spaces}

\phantomsection
\paragraph{Definition}
\label{sec:vector_spaces_definition} 
A \emph{vector space} $V$ is a non-empty set whose elements (called vectors) may be added together and multiplied by numbers (scalars), satisfying the following requirements (also called vector space axioms) for all vectors $\vv u, \vv v, \vv w \in V$ and scalars $a, b \in \BR$:\footnote{In general, vector spaces may be defined on any field $F$, not just the real numbers $F=\BR$ (e.g.\ the complex numbers $F=\mathbb{C}$ would also work). However, for our purposes, the restriction to real vector spaces is sufficient.}
\vspace{-\topsep}
\begin{enumerate}
    \item Vector addition is associative:\\
     $\vv u + (\vv v + \vv w) = (\vv u + \vv v) + \vv w $,\vspace{0.25\baselineskip}
    \item vector addition is commutative:\\
    $\vv u + \vv v = \vv v + \vv u$,\vspace{0.25\baselineskip}
    \item there exists a zero vector $\vv 0$:\\
    $\vv u + \vv 0 = \vv 0 + \vv u = \vv u$,\vspace{0.25\baselineskip}
    \item every vector $\vv u$ has an additive inverse $-\vv u$:\\
    $\vv u + (-\vv u) = (-\vv u) + \vv u = \vv 0$,\vspace{0.25\baselineskip}
    \item and the scalar multiplication satisfies:\\
    $a(\vv u + \vv v) = a\vv u + a\vv v$,\\
    $(a+b)\vv u = a\vv u + b\vv u$,\\
    $(ab)\vv u = a(b\vv u)$, and\\
    $1\vv u = \vv u$    
\end{enumerate}
\vspace{-\topsep}
(see e.g.\ Chapter~1 of Ref.~\citenum{roman2008advanced}).

For example, as suggested by the name, the familiar two--dimensional vectors in the plane or the three--dimensional vectors in space form the vector spaces $\BR^2$ and $\BR^3$. More generally, we have the vector spaces $\BR^n$ for any positive integer $n$. In fact, the ``vectors'' may be objects which we do not typically think of as vectors, e.g.\ $3\times 3$ matrices, or even functions (such as polynomials), as long as the vector space axioms listed above are satisfied.

It can be shown that any vector space has a \emph{basis}, i.e.\ a set of vectors $\vv e_i$ such that any vector can be written as a linear combination of these basis vectors in a unique way. While there are many different ways to choose such a basis set, it can be shown that the number of basis vectors is always the same, it is called the \emph{dimension} of the vector space $V$ and denoted $\dim(V)$. For example, the dimension of the vector space of $3 \times 3$ matrices is $9$\footnote{There are also vector spaces with infinite dimensions, for example, the vector space of all polynomials $\sum_i c_i x^i$ in one variable $x$ (a basis would be given by $\vv e_i \coloneqq x^i$). However, for our purposes only finite--dimensional vector spaces are relevant, e.g.\ the vector space of polynomials of a fixed degree.} (an explicit basis is given later in the text). Once a basis $\vv e_1,\dots, \vv e_n$ for an abstract $n$--dimensional vector space $V$ is chosen, we can write any vector $\vv v \in V$ as a linear combination of the basis vectors as
\begin{equation}
\vv v = \sum_{i=1}^{n} c_i \vv e_i\,.
\label{eq:abstract_vector_as_linear_combination}
\end{equation}
By ``collecting the values of the coefficients'' $c_i$, we can then identify the abstract vector $\vv v \in V$ with the concrete column vector $[c_1\ \cdots\ c_n]^\T\in \BR^n$ (an array of $n$ numbers).

\phantomsection
\paragraph{Linear maps}
\label{sec:linear functions} 
A function $f$ that maps vectors $\vv v \in V$ to the real numbers $\BR$ (or to another vector space) is called \emph{linear} if 
\begin{equation*}
  f(\vv v + \vv w) = f(\vv v) + f(\vv w) \quad \text{and}\quad f(a\vv v) = a f(\vv v)
\end{equation*}
for all vectors $\vv v, \vv w\in V$ and numbers $a\in \BR$. For example, taking the trace of a $3\times 3$ matrix, or the derivative of a polynomial, are both linear operations that can be written in this way.

If we identify abstract vectors in $V$ with concrete vectors in $\BR^n$, then linear functions $f:\BR^n\arrow \BR^m$ can be written as $m\times n$~matrices. Thus, any abstractly given vectors or linear maps can be ``translated'' to concrete column vectors or matrices. So in a way, the abstract notions of vector spaces and linear maps ``do not give anything new'' beyond arrays of numbers. However, when working with concrete arrays of numbers, the resulting formulas may appear quite complicated and obscure simple arguments and structures, so we choose to formulate the theory primarily in the more conceptual/abstract language.

\phantomsection
\subsection{Representations}
\label{sec:representations}
As mentioned at the end of the section on \hyperref[sec:groups]{groups}, the equivariant features used in \texttt{E3x} are elements of vector spaces, and groups such as $\SOthree$ act on these vector spaces by linear maps called \emph{representations}. We first give a definition of a representation in matrix form for concrete vector spaces~$\BR^n$, and then a more general definition of representations for arbitrary (abstract) vector spaces~$V$.
\phantomsection
\paragraph{Definition~1}
\label{sec:representations_definition1}
An $n$\nobreakdash--dimensional \emph{matrix representation} $\rho$ of a group $G$ on the vector space $\BR^n$ is a continuous function from $G$ to invertible square matrices $\in \BR^{n\times n}$ such that for all $g,h \in G$
\begin{equation*}
\rho(g)\rho(h) = \rho(g\odot h)
\end{equation*}
(see e.g.\ 0.3 in Ref.~\citenum{vinberg2010linear}).

In other words, the representation $\rho$ maps group elements to square matrices such that the group operation $\odot$ can be expressed by matrix multiplication. Similarly, the result of applying a transformation associated with a group element $g$ to an element $\vv{r}\in \BR^n$ can be expressed as $\rho(g)\vv{r}$. The vector space $\BR^n$ is also called the \emph{representation space}.

The groups $\SOthree$ and $\Othree$ are already given as sets of $3\times 3$~matrices, so they come with a ``canonical'' representation on $\BR^3$.\footnote{When $g\in\SOthree$ or $g\in\Othree$ is used in equations, this canonical representation, a $3\times 3$~matrix, is often meant implicitly. For example, we use this notation in \eqref{eq:MatrixConjugation}.} Also, every group has a ``trivial'' representation on $\BR$, which assigns the identity $1\times 1$~matrix to all group elements. There are also more interesting representations of $\SOthree$ and $\Othree$, which are introduced later in the text.

To avoid confusion, we want to stress that when talking about the ``dimension of a representation'', we always refer to the number $n$, the dimension of the representation space $\BR^n$. However, there are also other dimensions at play. For example, $n \times n$ matrices of course also form a vector space of dimension $n^2$. Further, the manifold $\rho(G)\subset \BR^{n\times n}$ of \emph{actually occurring} matrices $\rho(g)$ can have any dimension between $0$~and~$n^2$. For example, for the $n$--dimensional representation of any group $G$ that maps every $g\in G$ to the $n \times n$ identity matrix~$\mat I_n$, the manifold of actually occurring matrices has dimension~$0$ (all group elements are mapped to a ``single point'' in the space of $n \times n$ matrices). 

Before we get to the promised more interesting representations, we formulate the more general and abstract version of our definition.

\phantomsection
\paragraph{Definition~2}
\label{sec:representations_definition2}
A (linear, continuous, finite--dimensional)\footnote{Usually, the ``linear'' is implied when only ``representation'' is said, although there also exist ``projective'' representations. Similarly, when the group has a topological structure, also ``continuous'' is usually silently assumed. For general groups $G$, there are also important infinite--dimensional representations, but for compact groups $G$ they can all be assembled from finite--dimensional ones. So for compact groups, often ``finite--dimensional'' is also silently assumed.} representation of a group $G$ is a continuous function from $G$ to the invertible linear maps from a finite--dimensional vector space $V$ into itself, 
such that again for all $g,h \in G$
\begin{equation*}
\rho(g)\rho(h) = \rho(g\odot h)
\end{equation*}
(see e.g.\ 0.4 in Ref.~\citenum{vinberg2010linear}).

This is almost the same as \nameref{sec:representations_definition1}, but now instead of the concrete vector space $\BR^n$, we allow an $n$\nobreakdash--dimensional vector space $V$ that could be given in some other, more abstract way. Of course, we could find a basis of $V$ and write the maps $V\arrow V$ as matrices in this basis, which would lead us back to our first definition. However, the advantage of the second definition is that we can think about the concept of a representation without having to think about (or specify) a concrete basis, which may be extra work and only obscure the questions we are interested in.

When talking about representations, we may use the expressions ``a representation of $G$ on $V$'' or ``a $G$--representation on $V$'' interchangeably (they are synonymous). Further, if we talk about representations on different vector spaces $V$ and $W$ in the same context, we may refer to them as $\rho^V$ and $\rho^W$, respectively (to make clear which $\rho$ we are talking about). Finally, since a representation $\rho$ is always defined on some vector space $V$, when talking about the concept of a representation, the notation $(\rho, V)$ is most precise. However, it is also common in the literature to refer to $\rho$ or $V$ alone as ``representation'' if what is meant is clear from context (and we do so as well).

For us, the importance of representations is that they \emph{define} what an equivariant feature with values in a vector space is. Note that if we have an equivariant function $f: V\arrow W$, there is at most one permissible representation of $G$ on the span of $f(V)$, i.e.\ if the vector space is not ``unnecessarily large'' (meaning the image points of $f$ would only span a smaller subspace), the action of $G$ on $W$ is already determined by $f$. However, the fact that there \emph{must exist} a representation of $G$ on $W$ that makes $f$ equivariant is a strong restriction on $f$.

\phantomsection
\paragraph[Example~1]{Example 1: Matrices}
\label{sec:matrices_example1}
As a first example of a higher--dimensional representation of $G=\SOthree$, consider as representation space the vector space $V$ of $3\times 3$~matrices $\mat{M}$. Then, a representation $(\rho, V)$ is given by
\begin{equation}
   \rho(g) \mat{M} \coloneqq g \mat{M} g^{-1} = g \mat{M} g^\T 
   \label{eq:MatrixConjugation}
\end{equation}
for $g\in \SOthree$ (this is a representation of dimension~$9$).\footnote{Since all $g\in\SOthree$ are orthogonal matrices, inverting the matrix representing $g$ is the same as taking its transpose, i.e.\ $g^{-1} = g^\T$.} 

The notation used here may be confusing, so for clarity, we want to explain ``how to read''~\eqref{eq:MatrixConjugation} and how it relates to the two definitions given above. In this case, the representation is defined (in accordance with \nameref{sec:representations_definition2}) as the linear map  $\rho(g) \mat{M}$ that is computed as $g \mat{M} g^\T$, meaning ``$\mat{M}$ is (matrix) multiplied from the left with $g$ and from the right with $g^\T$'', where $g$ and $g^\T$ are the canonical representations ($3\times3$ matrices) of the group element $g$ and its transpose $g^\T$. To ``translate'' to \nameref{sec:representations_definition1}, we have to specify a basis $\vv e_1,\dots,\vv e_9$ for $V$. For example:
\begin{equation*}
\begin{aligned}
&\vv e_1 = \left[\begin{smallmatrix}
1 & 0 & 0\\
0 & 0 & 0\\
0 & 0 & 0\\
\end{smallmatrix}\right]\quad
\vv e_2 = \left[\begin{smallmatrix}
0 & 1 & 0\\
0 & 0 & 0\\
0 & 0 & 0\\
\end{smallmatrix}\right]\quad
\vv e_3 = \left[\begin{smallmatrix}
0 & 0 & 1\\
0 & 0 & 0\\
0 & 0 & 0\\
\end{smallmatrix}\right]\\
& \vv e_4 = \left[\begin{smallmatrix}
0 & 0 & 0\\
1 & 0 & 0\\
0 & 0 & 0\\
\end{smallmatrix}\right]\quad
\vv e_5 = \left[\begin{smallmatrix}
0 & 0 & 0\\
0 & 1 & 0\\
0 & 0 & 0\\
\end{smallmatrix}\right]\quad
\vv e_6 = \left[\begin{smallmatrix}
0 & 0 & 0\\
0 & 0 & 1\\
0 & 0 & 0\\
\end{smallmatrix}\right]\\
& \vv e_7 = \left[\begin{smallmatrix}
0 & 0 & 0\\
0 & 0 & 0\\
1 & 0 & 0\\
\end{smallmatrix}\right]\quad
\vv e_8 = \left[\begin{smallmatrix}
0 & 0 & 0\\
0 & 0 & 0\\
0 & 1 & 0\\
\end{smallmatrix}\right]\quad
\vv e_9 = \left[\begin{smallmatrix}
0 & 0 & 0\\
0 & 0 & 0\\
0 & 0 & 1\\
\end{smallmatrix}\right]\,.
\end{aligned}
\end{equation*}
Then, any (real) $3\times 3$ matrix can be written as
\begin{equation}
\mat{M} = \sum_{i=1}^{9} c_i \vv e_i\,,
\label{eq:matrix_written_as_vector}
\end{equation}
where the $c_i\in\BR$ are coefficients. Written in this form, it becomes clear that we could also specify $\mat M$ as a $9$--dimensional column vector containing the coefficients $c_i$ as entries, i.e.\ $[c_1 \ \cdots \ c_9]^\T\in\BR^9$. Then, the $9\times 9$ matrix representation of~$G$ (following \nameref{sec:representations_definition1}) can be chosen to ``have the same effect'' as  \eqref{eq:MatrixConjugation} on this vector, namely
\begin{equation*}
\begin{aligned}
\rho(g)&\coloneqq g \otimes g = \left[\begin{smallmatrix}
\varrho_1 & \varrho_2 & \varrho_3\\
\varrho_4 & \varrho_5 & \varrho_6\\
\varrho_7 & \varrho_8 & \varrho_9
\end{smallmatrix}\right] \otimes \left[\begin{smallmatrix}
\varrho_1 & \varrho_2 & \varrho_3\\
\varrho_4 & \varrho_5 & \varrho_6\\
\varrho_7 & \varrho_8 & \varrho_9
\end{smallmatrix}\right] \\&=  
\left[\begin{smallmatrix}
 \varrho_1 \varrho_1 & \varrho_2 \varrho_1 &  \varrho_3 \varrho_1 & \varrho_1 \varrho_2 &  \varrho_2 \varrho_2 & \varrho_3 \varrho_2 &  \varrho_1 \varrho_3 & \varrho_2 \varrho_3 &  \varrho_3 \varrho_3\\
 \varrho_4 \varrho_1 & \varrho_5 \varrho_1 &  \varrho_6 \varrho_1 & \varrho_4 \varrho_2 &  \varrho_5 \varrho_2 & \varrho_6 \varrho_2 &  \varrho_4 \varrho_3 & \varrho_5 \varrho_3 &  \varrho_6 \varrho_3\\
 \varrho_7 \varrho_1 & \varrho_8 \varrho_1 &  \varrho_9 \varrho_1 & \varrho_7 \varrho_2 &  \varrho_8 \varrho_2 & \varrho_9 \varrho_2 &  \varrho_7 \varrho_3 & \varrho_8 \varrho_3 &  \varrho_9 \varrho_3\\
 \varrho_1 \varrho_4 & \varrho_2 \varrho_4 &  \varrho_3 \varrho_4 & \varrho_1 \varrho_5 &  \varrho_2 \varrho_5 & \varrho_3 \varrho_5 &  \varrho_1 \varrho_6 & \varrho_2 \varrho_6 &  \varrho_3 \varrho_6\\
 \varrho_4 \varrho_4 & \varrho_5 \varrho_4 &  \varrho_6 \varrho_4 & \varrho_4 \varrho_5 &  \varrho_5 \varrho_5 & \varrho_6 \varrho_5 &  \varrho_4 \varrho_6 & \varrho_5 \varrho_6 &  \varrho_6 \varrho_6\\
 \varrho_7 \varrho_4 & \varrho_8 \varrho_4 &  \varrho_9 \varrho_4 & \varrho_7 \varrho_5 &  \varrho_8 \varrho_5 & \varrho_9 \varrho_5 &  \varrho_7 \varrho_6 & \varrho_8 \varrho_6 &  \varrho_9 \varrho_6\\
 \varrho_1 \varrho_7 & \varrho_2 \varrho_7 &  \varrho_3 \varrho_7 & \varrho_1 \varrho_8 &  \varrho_2 \varrho_8 & \varrho_3 \varrho_8 &  \varrho_1 \varrho_9 & \varrho_2 \varrho_9 &  \varrho_3 \varrho_9\\
 \varrho_4 \varrho_7 & \varrho_5 \varrho_7 &  \varrho_6 \varrho_7 & \varrho_4 \varrho_8 &  \varrho_5 \varrho_8 & \varrho_6 \varrho_8 &  \varrho_4 \varrho_9 & \varrho_5 \varrho_9 &  \varrho_6 \varrho_9\\
 \varrho_7 \varrho_7 & \varrho_8 \varrho_7 &  \varrho_9 \varrho_7 & \varrho_7 \varrho_8 &  \varrho_8 \varrho_8 & \varrho_9 \varrho_8 &  \varrho_7 \varrho_9 & \varrho_8 \varrho_9 &  \varrho_9 \varrho_9\\
\end{smallmatrix}\right]
\end{aligned}
\end{equation*}
where the $\varrho_1,\dots,\varrho_9$ are the entries of the canonical representation of $g$ (also a $3\times 3$ matrix), using the same basis as in \eqref{eq:matrix_written_as_vector}. This example also illustrates why it can be more convenient to work with \nameref{sec:representations_definition2} when talking about the theory (although for practical computations, \nameref{sec:representations_definition1} may be preferable).

\phantomsection
\paragraph[Example~2]{Example 2: Polynomials}
\label{sec:function_spaces_example2}
As a second example of a higher--dimensional representation of $G=\SOthree$, consider functions on the three--dimensional unit sphere $\BS^2 \subset \BR^3$. If we restrict them to functions that can be given as polynomials in $x,y,z$ (the Cartesian coordinates of a point on the sphere) with degree $\leq \ell$, this is a finite--dimensional vector space. Since the elements of $G$ map points on $\BS^2$ again into points of $\BS^2$, we can define a representation of $G$ on this space of functions $f:\BS^2\arrow \BR$ as
\begin{equation}
   \rho(g) f \coloneqq \Big( \vv r \mapsto f(g^{-1} \vv r) \Big)\,,
\label{eq:group_action_on_polynomials}
\end{equation}
where $\vv r = [x \ y \ z]^\T \in \BS^2$ is a point on the sphere and $g\in G$ is an element of the group.\footnote{It may not be immediately obvious why $g^{-1}$ (instead of $g$) has to be used on the right side in \eqref{eq:group_action_on_polynomials}: Recall from the definition of a representation at the start of this section that $\rho(g)\rho(h) = \rho(g \odot h)$ must be true for all $g, h \in G$. If $g^{-1}$ was replaced by $g$ in \eqref{eq:group_action_on_polynomials}, we would instead get $\rho(g)\rho(h) = \rho(h \odot g)$ ($\odot$ is not necessarily commutative, e.g.\ if $\odot$ represents matrix multiplication).}

This example is the basis of ``harmonic analysis on the sphere $\BS^2\subset \BR^3$'', which is a higher--dimensional analogue to the Fourier analysis of periodic functions (i.e.\ functions on the circle $\BS^1$). It can be applied e.g.\ to the study of polynomials on $\BR^3$: A polynomial on $\BR^3$ decomposes into a sum of homogeneous polynomials,\footnote{Homogeneous polynomials only contain non-zero terms of the same degree, e.g.\ \mbox{$x^2 + 2xy$} is homogeneous, but \mbox{$3y^2 + x$} is not.} which are already determined by their values on~$\BS^2$.

\phantomsection
\paragraph{Invariant subspaces}
\label{sec:invariant_subspaces}
If we have a representation $(\rho, V)$ of a group $G$ and a subspace $W \subseteq V$ that is mapped to itself by all $\rho(g)$, we call $W$ an \emph{invariant subspace}. In this case, the restriction of the linear map \mbox{$\rho(g): V\arrow V$} to the subspace $W$ gives again a representation (on $W$).

To give an example for invariant subspaces, consider the $9$--dimensional $\SOthree$--representation on the space  of $3 \times 3$ matrices discussed in \nameref{sec:matrices_example1}. In \eqref{eq:MatrixConjugation}, the symmetric and anti--symmetric matrices form such invariant subspaces, because a symmetric matrix $A=A^\T$ is mapped by $\rho(g), g\in SO(3)$ to $gAg^\T=(gA^\T g^\T)^\T = (g A g^\T)^\T$ and similarly for an anti--symmetric matrix $A=-A^\T$.
Since symmetric matrices can be written as e.g.\
\begin{equation*}
\begin{aligned}
\mat S = \ 
&c_1 \left[\begin{smallmatrix} 
1 & 0 & 0\\
0 & 0 & 0\\
0 & 0 & 0
\end{smallmatrix}\right] + 
c_2 \left[\begin{smallmatrix} 
0 & 0 & 0\\
0 & 1 & 0\\
0 & 0 & 0
\end{smallmatrix}\right] + 
c_3 \left[\begin{smallmatrix} 
0 & 0 & 0\\
0 & 0 & 0\\
0 & 0 & 1
\end{smallmatrix}\right]\\ &\quad+
 c_4 \left[\begin{smallmatrix} 
0 & 1 & 0\\
1 & 0 & 0\\
0 & 0 & 0
\end{smallmatrix}\right]
+
c_5 \left[\begin{smallmatrix} 
0 & 0 & 1\\
0 & 0 & 0\\
1 & 0 & 0
\end{smallmatrix}\right] 
+
c_6 \left[\begin{smallmatrix} 
0 & 0 & 0\\
0 & 0 & 1\\
0 & 1 & 0
\end{smallmatrix}\right]\,,
\end{aligned}
\end{equation*}
and anti--symmetric matrices as e.g.\
\begin{equation*}
\mat A = \ 
c_1 \left[\begin{smallmatrix} 
0 & 0 & 0\\
0 & 0 & 1\\
0 & -1 & 0
\end{smallmatrix}\right] +
c_2 \left[\begin{smallmatrix} 
0 & 0 & -1\\
0 & 0 & 0\\
1 & 0 & 0
\end{smallmatrix}\right] +
c_3 \left[\begin{smallmatrix} 
0 & 1 & 0\\
-1 & 0 & 0\\
0 & 0 & 0
\end{smallmatrix}\right] 
\,,
\end{equation*}
respectively, they are representations of dimensions~$6$~and~$3$. Further, also the subspace of traceless symmetric matrices is an invariant subspace, which gives a representation of dimension~$5$.\footnote{For a symmetric matrix to have trace~0, it must be that \mbox{$c_3=-(c_1+c_2)$}, hence the subspace has only dimension $5$ (only five $c_i$ can be chosen freely). However, giving an explicit basis is slightly more involved than for symmetric and anti--symmetric matrices and is therefore postponed until later in the text.}

\phantomsection
\paragraph{Isomorphisms}
\label{sec:isomorphisms}
We can ask when two representations are ``essentially the same'', i.e.\ differ only ``by the name of the vectors'' (meaning the elements of the representation space). This is captured in the definition of an isomorphism between $G$\nobreakdash--representations $(\rho^V, V)$ and $(\rho^W, W)$. The representations $(\rho^V, V)$ and $(\rho^W, W)$ are called \emph{isomorphic} if there is a linear bijection\footnote{A bijective mapping $f:A\arrow B$ between two sets $A$ and $B$ means that there is a one-to-one correspondence between elements of $A$ and elements of $B$, i.e.\ every element $b\in B$ is the image of exactly one element $a\in A$.} $f: V \rightarrow W$ between the vector spaces $V$ and $W$ such that
\begin{equation*}
    f(\rho^V(g) \vv v) = \rho^W(g) f(\vv v)
\end{equation*}
for all $g\in G, \vv v\in V, f(\vv v) \in W$. If a linear bijection $f: V \rightarrow W$ with this property exists, we write $(\rho^V, V) \simeq (\rho^W, W)$, meaning $(\rho^V, V)$ and 
$(\rho^W, W)$ are ``equal up to isomorphism''.

Of course, two isomorphic representations must have the same dimension, but the converse is not true: The canonical representation of $\SOthree$ on $\BR^3$ is \emph{not} isomorphic to the one that always assigns the $3\times 3$ identity matrix, although both use the same $3$--dimensional representation space $\BR^3$. However, we can say that the $3$\nobreakdash--dimensional representation on anti--symmetric $3\times 3$ matrices discussed before is isomorphic to the canonical representation: An isomorphism is given by
\begin{equation}
   \begin{bmatrix} 0 &  z & -y\\
                  -z &  0 &  x \\
                   y & -x & 0
  \end{bmatrix}
  \mapsto
  \begin{bmatrix} x \\ y\\ z \end{bmatrix}\,.
\label{eq:antisymm_3x3_matrix}
\end{equation}

Also, the representation on all functions on the sphere $\BS^2\subset \BR^3$ that can be given by homogeneous polynomials of degree~$2$ in $x,y,z$ (see \nameref{sec:function_spaces_example2}) is isomorphic to the $6$\nobreakdash--dimensional representation on symmetric $3\times 3$~matrices. An isomorphism is given by
\begin{equation}
  \mat S \mapsto \Big(\vv r \mapsto \vv r^\T \mat S \vv r\Big)\,,
  \label{eq:quadratic_forms}
\end{equation}
where $\vv r = [x \ y \ z]^\T$. This means that the symmetric matrix
\begin{equation*}
\mat S =
\begin{bmatrix}
c_1 & c_4 & c_5 \\
c_4 & c_2 & c_6 \\
c_5 & c_6 & c_3
\end{bmatrix}
\end{equation*}
is mapped to the polynomial
\begin{equation*}
\begin{aligned}
  f(x, y, z) =\ &c_1 \cdot x^2 + c_2 \cdot y^2 + c_3 \cdot z^2\\
  &\quad+ 2 c_4 \cdot xy + 2 c_5 \cdot xz + 2 c_6 \cdot yz\,.
\end{aligned}
\end{equation*}

\phantomsection
\subsection[Decomposition into irreps]{Decomposition into irreducible components}
\label{sec:irreps}

Given two matrix representations (\nameref{sec:representations_definition1}) $\rho^{W_1}$ on \mbox{$W_1=\BR^m$} and $\rho^{W_2}$ on \mbox{$W_2=\BR^n$}, the combined representation on the concatenated tuples of numbers \mbox{$V=\BR^{m+n}$} is called a \emph{direct sum} representation, which we write as $V = W_1 \oplus W_2$. The corresponding $\rho^{V}(g)$ are block diagonal matrices composed of an $m\times m$~block $\rho^{W_1}(g)$ and an $n\times n$~block $\rho^{W_2}(g)$.

More generally, for a linear representation on~$V$ (\nameref{sec:representations_definition2}) and two invariant subspaces $W_1, W_2 \subseteq V$, we say $V$ is the \emph{direct sum} of $W_1$ and $W_2$ if every vector $\vv v \in V$ can be uniquely written as $\vv v = \vv w_1 + \vv w_2$ with $\vv w_1\in W_1$ and $\vv w_2\in W_2$. As a consequence, a basis $\vv e_1,\dots, \vv e_m$ of $W_1$ and $\vv e_{m+1},\dots,\vv e_{m+n}$ of $W_2$ then give a basis $\vv e_1,\dots,\vv e_{m+n}$ of $V$, leading back to the definition for matrix representations (see above). This can be generalized in the obvious way to direct sums of more than two subrepresentations.

\phantomsection
\paragraph{Definition} 
\label{sec:irrep_definition}
A representation on $V$ that has no invariant subspaces except $\{\vv 0\}$\footnote{Since the subspace $\{\vv 0\}$ (that contains only the zero vector $\vv 0$) is always mapped to itself by any linear map, $\{\vv 0\}$ is always a ``trivial'' subspace of any vector space $V$.} and $V$ itself is called \emph{irreducible}. In particular, \emph{irreducible representations} (irreps) cannot be written as a direct sum of smaller representations.

For example, recall that for the $9$--dimensional $\SOthree$--representation on the space of $3\times 3$ matrices (see \nameref{sec:matrices_example1}), symmetric and anti--symmetric matrices form invariant subspaces. Thus, the representation can be written as the direct sum of the $6$--dimensional and $3$--dimensional subrepresentations of symmetric and anti--symmetric matrices. Further, the symmetric matrices can be written as a direct sum of the $1$\nobreakdash--dimensional subspace $\BR\cdot \mat I_3$ ($ \mat I_3$ denotes the $3\times3$ identity matrix) and the $5$\nobreakdash--dimensional subspace of traceless symmetric matrices. Together, this gives that the $9$--dimensional representation in $\eqref{eq:MatrixConjugation}$ is the direct sum of (irreducible) representations of dimensions $1$, $3$, and $5$. 

One can show that for ``compact groups'' (which include $\SOthree$ and $\Othree$, but not $\SEthree$ and $\Ethree$, see Table~\ref{tab:groups_relevant_in_3d}), every linear, finite\nobreakdash--dimensional representation can be written as a direct sum of irreducible representations (see e.g.\ Theorem 2 in chapter 2.5 of Ref.~\citenum{vinberg2010linear} and its corollary).

\phantomsection
\paragraph[\texorpdfstring{$\SOthree$}{SO(3)}--equivariant features]{$\SOthree$--equivariant features}
\label{sec:so3_equivariant}
Recall that our goal is to describe equivariant features, and we said that our features $f:\BR^3\arrow V$ take values in a vector space $V$ on which e.g.\ the group $\SOthree$ acts as a linear representation. Similar to the construction we used for $\SOtwo$--equivariant functions (see \nameref{sec:equivariance}), we can again reduce this to studying $\SOthree$--equivariant functions on the three--dimensional unit sphere $b_1,\dots,b_n:\BS^2\arrow V$, and combining them with radial functions $a_1,\dots,a_n: \BR_{\geq 0} \arrow \BR$ (with $a_i(0)=0$). Thus, all $\SOthree$--equivariant functions can be written as
\begin{equation}
f(\vv r) \coloneqq \sum_{i=1}^n a_i(r)\cdot b_i(\hat{\vv{r}})\,,
\label{eq:radial_so3}
\end{equation}
where $\vv r \in \BR^3$, $r = \lVert \vv r \rVert$, and $\hat{\vv{r}} = \vv r / r$.\footnote{An $\SOthree$--equivariant function on the sphere $\BS^2$ is already determined by its value at one point, e.g. at $[1\ 0\ 0]^\T$, since for every other point $\vv u\in \BS^2$ there is a rotation $g\in \SOthree$ that moves $[1\ 0\ 0]^\T$ to $\vv u$. The $\SOthree$--equivariant functions $\BS^2\subset \BR^3$ form a vector space (they are closed under addition and scalar multiplication), so also the possible values at $[1\ 0\ 0]^\T$ form a vector space, which is a subspace of $V$ and has as dimension of at most $\dim(V)$. So if we let $b_1,\dots,b_n$ be a basis of this vector space of equivariant functions $\BS^2\arrow V$, we can  write any equivariant function $f:\BR^3\arrow V$ uniquely in the form \eqref{eq:radial_so3}.
To construct a specific $f(\vv r)$ with some desired behaviour, it is enough to determine the corresponding radial functions $a_1,\dots,a_n:\BR_{\geq 0}\arrow \BR$ in one specific direction (e.g.\ on the positive $x$--axis) and $f(\vv r)$ automatically \emph{must have} the desired behavior in all directions by equivariance (compare to the argument presented at the end of \nameref{sec:equivariance}).}

In the case of $G=\SOtwo$ and the canonical representation on $\BR^2$, the functions $n$ and $t$, giving vectors in radial/normal and in tangential direction to the circle, are a basis for equivariant functions $f:\BS^1\arrow \BR^2$. In three dimensions, however, while we can specify vectors in the radial direction, there are no vectors in the tangential direction that could be values of an equivariant function $f:\BS^2\arrow \BR^3$. To see why this is the case, consider that there is just one possible tangential direction for each point on the circle $\BS^1$, but there are infinitely many possible tangential directions for each point on the sphere $\BS^2$. It might be tempting to just pick one of these directions, but this would violate equivariance as follows: Pick rotations around a specific axis, e.g.\ the $x$--axis. Since rotations around the $x$--axis leave the point $[1\ 0\ 0]^\T$ fix, also the value of any equivariant function $\BS^2\arrow V$ at $[1\ 0\ 0]^\T$ must remain fixed under $\rho(g)$ for all rotations $g$ around the $x$--axis. But any tangential direction we could pick at the point $[1\ 0\ 0]^\T$ would necessarily rotate when we rotate around the $x$--axis, thus it cannot be a function value for an equivariant function.

If a representation on $V$ is reducible, i.e.\ \mbox{$V\simeq V_1 \oplus \cdots \oplus V_k$}, then equivariant features with values in $V$ contain no ``new information'' that is not already contained in equivariant features with values in the smaller vector spaces $V_1,\dots,V_k$. Therefore, it is sufficient to focus on irreducible representations $(\rho,V)$ of $\SOthree$. It turns out that there exists only one irreducible representation of $\SOthree$ on $V$ (up to isomorphism) for each dimension $1,3,5,\dots$ and only one equivariant map $\BS^2\arrow V$ (up to scalar multiples), which is given by the so-called spherical harmonics $Y_\ell:\BS^2\arrow \BR^{2\ell+1}$, which are discussed in detail later in the text. This means that all information an equivariant feature could contain is already encoded in features of the form $a(r) \cdot Y_\ell(\hat{\vv{r}})$, which are the basis of the geometric features in \texttt{E3x}.

\phantomsection
\paragraph[Irreps of \texorpdfstring{$\SOthree$}{SO(3)}]{Irreducible representations of $\SOthree$}
\label{sec:irreps_of_so3}

As stated above, irreducible representations of $\SOthree$ on $\BR^n$ exist only for odd dimensions $n=2\ell+1$, where $\ell$ can be any positive integer or zero and is called the \emph{degree} of the representation. These occur naturally on $\BS^2$: Consider the vector space $P^\ell$ of homogeneous polynomials of degree $\ell$ in variables $x$, $y$, and $z$ as a representation of $\SOthree$ with the group action given by \eqref{eq:group_action_on_polynomials}. The polynomial $r^2 = x^2+y^2+z^2$ is identically~$1$ on the unit sphere $\BS^2$, so multiplication with $r^2$ gives an equivariant map $P^\ell\arrow P^{\ell+2}$. There is a natural scalar product on these vector spaces of polynomials that is invariant under rotations, and under which the adjoint of the operators ``multiplying with $x$, $y$, $z$'' are the operators $\frac{\partial}{\partial x}$, $\frac{\partial}{\partial y}$, $\frac{\partial}{\partial z}$.\footnote{This means that e.g.\ $\langle xf, g\rangle = \langle f, \frac{\partial}{\partial x} g\rangle$, see Ref.~\citenum{vinberg2010linear}, section~III.9.3 for details.} It follows that the adjoint operator of the multiplication with $r^2 = x^2+y^2+z^2$ is the Laplace operator
\begin{equation*}
   \Delta = \frac{\partial^2}{\partial x^2} + \frac{\partial^2}{\partial y^2} + \frac{\partial^2}{\partial z^2} : P^{\ell+2}\arrow P^\ell,
\end{equation*}
which is then also invariant under rotations. Thus, we can decompose the $\SOthree$--representations $P^\ell$ 
for $\ell\geq 2$ as a direct sum
\begin{equation*}
     P^\ell = H^\ell \,\oplus\, r^2\cdot P^{\ell-2}
\end{equation*}
for $H^\ell \coloneqq \{f\in P^\ell\,|\, \Delta f = 0\}$ ($H^\ell$ is the subspace of $P^\ell$ that contains only \emph{harmonic} functions, i.e.\ satisying $\Delta f = 0$). One can then show that this $H^\ell$ is a representation of $\SOthree$ of dimension $2\ell+1$, it is irreducible, and \emph{every} irreducible representation of $\SOthree$ is isomorphic to \emph{exactly one} of these (see e.g.\ Chapter~7 in Ref.~\citenum{kosmann2009}).

\phantomsection
\paragraph{Dual representation}
\label{sec:dual}
For any representation $(\rho, V)$ of $G$ there is a representation on the dual space $\check V$ of linear functions $L:V\arrow \BR$, which is given by the usual formula for functions
\begin{equation*}
    \check\rho(g) L \coloneqq \Big( \vv v \mapsto L\big(\rho(g^{-1}) \vv v\big) \Big)\,.
\end{equation*}
One can show (see e.g.\ Ref.~\citenum{vinberg2010linear}, Theorem~2 in chapter~I.2) that for compact groups like $\SOthree$ or $\Othree$, there exists for any representation $(\rho, V)$ of~$G$ a $G$--invariant scalar product on $V$. This means in particular that
\begin{equation*}
   \langle \vv v, \vv w\rangle =   \langle\rho(g) \vv v, \rho(g)\vv w\rangle = 
   \langle \vv v, \rho(g)^\T\rho(g)\vv w\rangle
\end{equation*}
for all $\vv v, \vv w\in V$, which can only be the case if $\rho(g)\rho(g)^\T$ is the identity matrix, i.e.\ the $\rho(g)$ must be orthogonal matrices.

In those cases (in particular, always when $G$ is compact) the dual representation $(\check \rho, \check V)$ is isomorphic to $(\rho, V)$. An isomorphism is given by
\begin{equation*}
   V\arrow \check V,\quad \vv v \mapsto \Big(\vv w \mapsto \langle \vv v, \vv w\rangle\Big)\,.
\end{equation*}
This means that for compact groups, we do not have to distinguish between a representation and its dual, since they are isomorphic (except for giving explicit formulas, when we have to specify the scalar product and apply the isomorphism).

\phantomsection
\paragraph{Spherical harmonics}
\label{sec:spherical_harmonics}

Consider the action of $G=\SOthree$ on the sphere $\BS^2$, and our representation on $H^\ell$, which are functions on $\BS^2$. Then we can map each point $\vv u \in \BS^2$ to a linear form on $H^\ell$ by just evaluating the function $f:\BS^2\arrow \BR$ at the point $\vv u$. This gives an equivariant map $\BS^2 \arrow \check H^\ell$. Since this dual space is isomorphic to the original function space, it also gives an equivariant map $\BS^2 \arrow H^\ell$.

To make this map and the representation on $H^\ell$ explicit, we consider as scalar product of functions $f_1,f_2:\BS^2\arrow \BR$ the surface integral
\begin{equation*}
   \langle f_1, f_2 \rangle \coloneqq \int_{\BS^2} f_1(\vv u) f_2(\vv u) d\vv u\,.
\end{equation*}
This is an $\SOthree$--invariant scalar product, and we now have all ingredients to give an orthonormal basis of $H^{\ell}$. Since we are interested in points \mbox{$\vv u \in \BS^2$} on the sphere and $H^\ell$ only contains harmonic functions, these basis functions are also called \emph{spherical harmonics} and denoted $Y_\ell^m$ (the meaning of $m$ will be explained below).

\begin{figure}[tbp]
    \centering
    \includegraphics[width=\columnwidth]{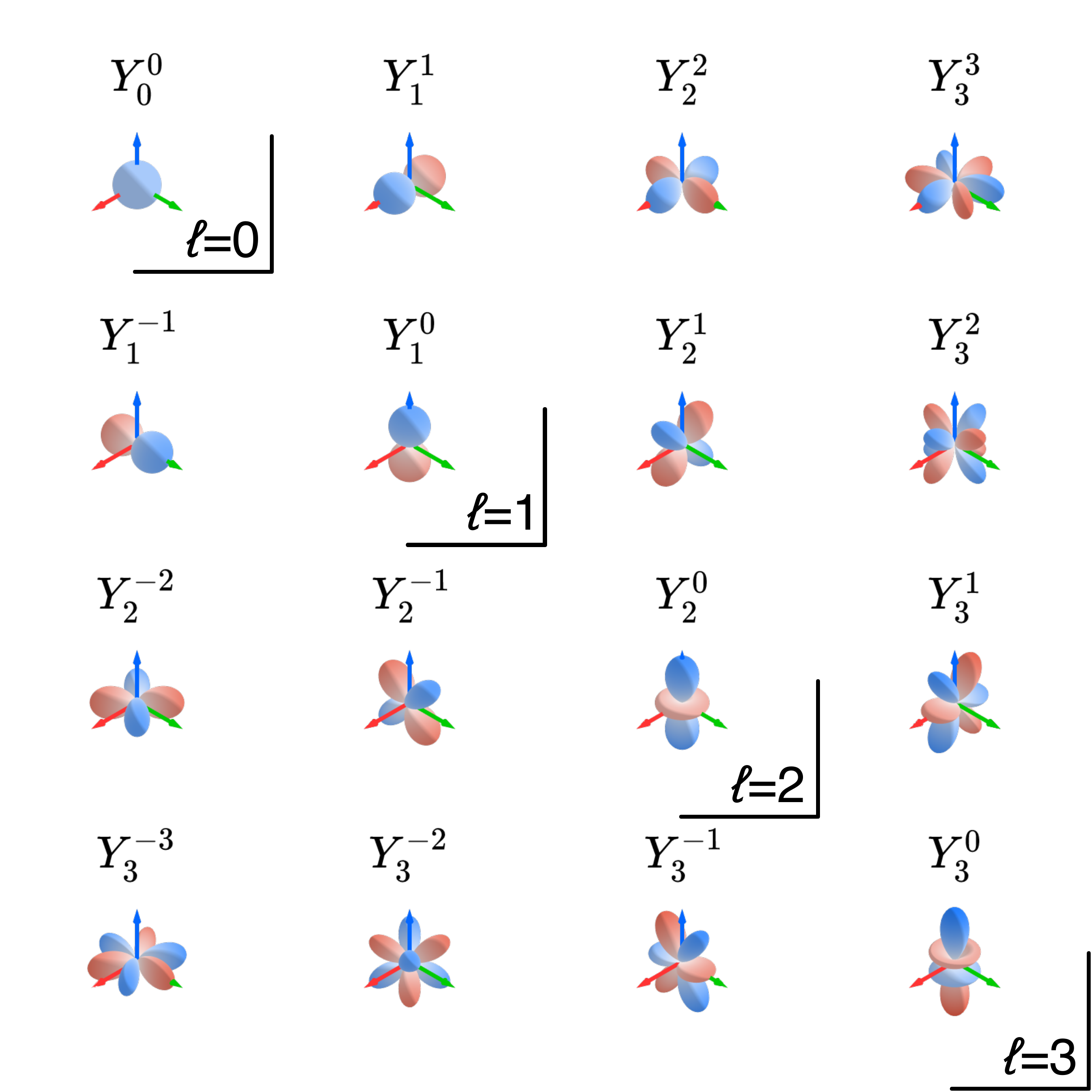}
    \caption{Visualization of the spherical harmonics $Y_{\ell}^{m}$ up to degree $\ell=3$ (positive values shown in blue, negative values in red, the $x$--, $y$--, and $z$--axes are depicted as red, green, and blue arrows). All $Y_{\ell}^{m}$ are evaluated on the unit sphere, but its radius at point $\vv r$ is scaled by $|Y_{\ell}^{m}(\vv r)|$, leading to distinct shapes for the different $Y_{\ell}^{m}$.}
    \label{fig:spherical_harmonics}
\end{figure}

For $\ell=0$, this space consists only of the constants, so only one orthonormal basis is possible. For the $(2\ell+1)$--dimensional spaces with $\ell>0$, there are more choices (different conventions for the spherical harmonics differ by choosing slightly different orthonormal bases). For the convention used in \texttt{E3x}, it is useful to consider the action of the subgroup $\mathrm{SO}(2)$ of rotations around the $z$--axis on $H^\ell$. One can show that although the $H^\ell$ are irreducible as $\SOthree$--representations, they decompose into $\ell+1$ different representations of $\mathrm{SO}(2)$: A one--dimensional subspace that is the trivial representation, i.e.\ it remains fixed under rotations, and $\ell$ two--dimensional subspaces in which the rotations by angle $\alpha$ act as rotations by angle $m\cdot \alpha$ for $m=1,2,\dots,\ell$. We choose an orthonormal basis of $H^{\ell}$ such that $Y^0_\ell$ remains fixed under rotations around the $z$--axis, and the $Y^{\pm m}_\ell$ span the subspace corresponding to the $m$ above (see Fig.~\ref{fig:spherical_harmonics}).

So far, we have considered the $Y_\ell^m$ only as functions on $\BS^2$, but it is custom to extend them to $\BR^3\setminus\{\vv 0\}$ such that they depend only on $\hat{\vv{r}} = \vv r/r$ with $r = \lVert\vv r\rVert$. Then, the general formulas are:

\begin{strip}
\begin{equation}
\begin{aligned}
&Y_{\ell}^{m}(\vv{r}) = \sqrt{\frac{2\ell+1}{4\pi}}\begin{cases}
    \sqrt{2}\cdot \Pi_\ell^{\lvert m\rvert}(z) \cdot \Im\Big((x+iy)^{|m|}\Big) 
    & m < 0 \\
    \Pi_\ell^{0}(z)
    & m = 0 \\
    \sqrt{2}\cdot \Pi_\ell^{m}(z) \cdot \Re\Big((x+iy)^m\Big)
    & m > 0 \\
    \end{cases}
\\
&\Pi_{\ell}^{m}(z) = \sqrt{\frac{(\ell-m)!}{(\ell+m)!}}
    \sum_{k=0}^{\lfloor\frac{\ell-m}{2}\rfloor} \ \frac{(-1)^k}{2^\ell}
    \binom{\ell}{k} \binom{2\ell-2k}{\ell}\frac{(\ell-2k)!}{(\ell-2k-m)!}
      r^{2k}z^{\ell-2k-m}\\
&\Im\Big((x+iy)^{m}\Big) = \sum_{k=0}^{ m}\binom{m}{k}x^{k} y^{m-k} \sin\left(\frac{\pi}{2}(m-k)\right) \quad (\mathrm{for}\ m > 0) 
\\
&\Re\Big((x+iy)^m\Big) = \sum_{k=0}^{m}\binom{m}{k}x^{k} y^{m-k} \cos\left(\frac{\pi}{2}(m-k)\right) \quad (\mathrm{for}\ m > 0) 
\end{aligned}
\label{eq:spherical_harmonics}
\end{equation}
\end{strip}
where $x$, $y$, and $z$ are the entries of $\vv{r}=[x \ y \ z]^\T$ and $i$ is the imaginary unit. The \emph{order} $m$ (see above) can take any integer value between $-\ell$ and $\ell$, i.e.\ there are $(2\ell+1)$ valid orders $m$ per degree $\ell$. Although the general definition of spherical harmonics given in \eqref{eq:spherical_harmonics} may look quite involved, for particular values of $\ell$ and $m$, the expression for $Y_\ell^m$ can be strongly simplified. For example, the first few spherical harmonics are
\begin{equation*}
\begin{aligned}
Y_{0}^{0}&=\sqrt{\frac{1}{4\pi}} \\
Y_{1}^{1}=\sqrt{\frac{3}{4\pi}}\frac{x}{r} \quad
Y_{1}^{-1}&=\sqrt{\frac{3}{4\pi}}\frac{y}{r} \quad
Y_{1}^{0}=\sqrt{\frac{3}{4\pi}}\frac{z}{r} \,.
\end{aligned}
\end{equation*}
Since these $Y_l^m$ are an orthonormal basis, they also give the equivariant map $\BS^2 \arrow H^\ell$ by
\begin{equation*}
   \vv r \mapsto \sum_{m=-\ell}^\ell Y_\ell^m(\vv r) \cdot Y_\ell^m \in H^\ell\,.
\end{equation*}
The use of both $Y_\ell^m(\vv r)$ and $Y_\ell^m$ in the same context may be slightly confusing: Here, $Y_\ell^m(\vv r)$ is the function $Y_\ell^m$ evaluated at the point $\vv r$ (giving a scalar coefficient), which is multiplied with the function $Y_\ell^m$ itself (a polynomial of degree $\ell$ in $x$, $y$, and $z$), which is used as a basis vector of the ($2\ell+1$)--dimensional vector space $H^\ell$. In other words, expressed in the basis $Y_\ell^\ell, Y_\ell^{-\ell}, \dots, Y_\ell^{0}$ we map a vector $\vv r\in \BR^3$ to the vector $[Y_\ell^{\ell}(\vv r) \ Y_\ell^{-\ell}(\vv r) \ \cdots \ Y_\ell^{0}(\vv r)]^\T \in \BR^{2\ell+1}$ containing the values of the coefficients for each basis function (compare to the way we wrote an abstract vector $\vv v$ as a linear combination of basis vectors in \eqref{eq:abstract_vector_as_linear_combination}). Note that the order of the basis functions is irrelevant (different conventions are possible), as long as a consistent choice is made. In \texttt{E3x}, the default ordering convention is $Y_{\ell}^{\ell}, Y_{\ell}^{-\ell}, Y_{\ell}^{\ell-1}, Y_{\ell}^{-\ell+1}, \dots,  Y_{\ell}^{0}$, because for $\ell=1$, this corresponds to the usual $(x,y,z)$--order for three--dimensional vectors.


\phantomsection
\paragraph{Schur's lemma}
\label{sec:Schur}
We have already introduced the notion of \hyperref[sec:isomorphisms]{isomorphisms} and said two representations are isomorphic if there exists an isomorphism between them. If there is one isomorphism $f:V\arrow W$, there are also infinitely many others: Any scalar multiple $\lambda f$ with $\lambda \neq 0$ is also an isomorphism. Schur's lemma says these are already \emph{all} the isomorphisms if $(\rho, V)$ and $(\rho, W)$ are irreducible representations of $G=\SOthree$ or $G=\Othree$.

Although straightforward to prove,\footnote{Typically this is formulated and proved over the complex numbers
(see e.g.\ Ref.\citenum{vinberg2010linear}, I.4.2), but the usual proof also works for real vector spaces of odd dimensions, since there must be at least one real eigenvalue.} this is a surprising statement: In general, there are many more choices for an isomorphism. For example, for vector spaces with trivial group operation, isomorphisms would be linear invertible maps, and there are ``a lot'' of isomorphisms of vector spaces $\BR^n\arrow\BR^n$ (every possible invertible $n\times n$ matrix corresponds to one such map).
The reason Schur's lemma is true for $\SOthree$ and $\Othree$ is that it is true for irreducible representations on real vector spaces of odd dimension, and for $\SOthree$ and $\Othree$, \emph{all} irreducible representations happen to have odd dimension.

We also said that for $\SOthree$ and $\Othree$, any two irreducible representations of the same dimension are isomorphic, i.e.\ there exists an isomorphism $f: V \arrow W$. Thus Schur's lemma allows defining a concrete map between two irreducible representations of the same dimension that are abstractly given in two different ways, just by fixing the scalar factor. On irreducible representations of $\SOthree$ or $\Othree$ there is also a $G$--invariant scalar product (see e.g.\ Ref.~\citenum{vinberg2010linear}, I.2.5), and one can use Schur's lemma to show that it is also unique up to a scalar factor. So if we have in addition specified $G$--invariant scalar products on both vector spaces $V$ and $W$, there is an isomorphism that also respects the scalar product, and this isomorphism is unique up to a factor $\pm 1$ (this will be important in the next section).

\phantomsection
\subsection[Coupling irreps]{Coupling irreps with tensor products}
\label{sec:coupling_irreps}

\phantomsection
\paragraph{Definition}
\label{sec:tensor_product_definition}
Given representations $\rho^V$ and $\rho^W$ of a group $G$ on vector spaces $V$ and $W$, there also exists a representation $\rho^{V \otimes W}$ on the tensor product $V \otimes W$ of the vector spaces. It has the following properties:
\vspace{-\topsep}
\begin{enumerate}
\item There is a map $V\times W \arrow V \otimes W$, it is linear in each variable, and written as $\vv v,\vv w\mapsto \vv v\otimes \vv w$.\vspace{0.25\baselineskip}
\item If $\vv e_i$ are basis vectors of $V$, and $\vv f_j$ basis vectors of $W$, then the $\vv e_i\otimes \vv f_j$ are a basis of $V \otimes W$.\vspace{0.25\baselineskip}
\item The representation $\rho^{V \otimes W}$ on $V \otimes W$ has the property
\begin{equation*}
 \rho^{V \otimes W}(g)(\vv v\otimes \vv w) = \rho^V(g)(\vv v) \otimes \rho^W(g)(\vv w)\,.
\end{equation*}
\end{enumerate}
\vspace{-\topsep}
It can be shown that a representation with these properties is unique up to isomorphism (note that the tensor product can be defined in different equivalent ways, here we follow \mbox{Ref.~\citenum{serre2012linear}}, chapter~1.5).

One way of realizing the tensor product of $V=\BR^m$ and $W=\BR^n$ in a concrete way is as $m\times n$~matrices: The bilinear map 
\mbox{$V\times W \arrow V \otimes W$} can be taken as $\vv v,\vv w \mapsto \vv v \vv w^\T$, and the representation $\rho(g)$ of a group element $g$ on the $m\times n$ matrices $\mat M = \vv v \vv w^\T$ can be defined as
\begin{equation*}
    \rho^{V\otimes W}(g)\mat M \coloneqq \rho^V(g) \mat M \rho^W(g)^\T\,.
\end{equation*}

This tensor product representation is typically not irreducible, but can be re-written as a direct sum of irreps (see \nameref{sec:irreps}). Thus, two irreps can be ``coupled'' via a tensor product to produce new irreps.

We have seen this already for the canonical representation $\BR^3$ of $\SOthree$: The vector space $\BR^3 \otimes \BR^3$ of $3\times 3$~matrices decomposes into $1$--, $3$--, and $5$\nobreakdash--dimensional representations. The Clebsch--Gordan theorem (see e.g.\ Ref.~\citenum{kosmann2009}, chapter 6, section 2.2) generalizes this to tensor products of $(2\ell_1+1)$\nobreakdash--dimensional and $(2\ell_2+1)$\nobreakdash--dimensional irreps.

For ease of notation, we write the irreducible $\SOthree$--representations of degree $0,1,2,...$ as $\irreps{0}, \irreps{1}, \irreps{2},...$. The general Clebsch--Gordan rule for expressing the tensor product $\irreps{a} \otimes \irreps{b}$ of two irreps $\irreps{a}$ and $\irreps{b}$ of $\SOthree$ as a direct sum is
\begin{equation*}
 \irreps{a} \otimes \irreps{b} \simeq
    \irreps{\left(\lvert a-b \rvert\right)}  \oplus
   \dots \oplus
    \irreps{\left(a+b \right)}.
\end{equation*}

For example, the tensor product of two vectors (irrep $\irreps{1}$) can be written as
\begin{equation*}
\irreps{1}\otimes\irreps{1} \simeq \irreps{0}\oplus\irreps{1}\oplus\irreps{2}\,.
\end{equation*}
To simplify notation, we introduce the short-hand `$\couples{\ell}$' to refer to the irrep of degree~$\ell$ in the direct sum representation of a tensor product, i.e.\
\begin{equation*}
\irreps{1} \otimes \irreps{1} 
=
\overbrace{\left(\irreps{1}\couples{0}\irreps{1}\right)}^{\simeq\irreps{0}} \oplus
\overbrace{\left(\irreps{1}\couples{1}\irreps{1}\right)}^{\simeq\irreps{1}} \oplus
\overbrace{\left(\irreps{1}\couples{2}\irreps{1}\right)}^{\simeq\irreps{2}}\,.
\end{equation*}

To get a concrete formula for this ``coupling'', we need to specify one particular isomorphism from the subspace  $\irreps{a}\couples{c}\irreps{b}$ to $\irreps{c}$. By \nameref{sec:Schur}, we can fix it up to a factor $\pm 1$ by specifying scalar products (or equivalently: norms) on both sides. On our concrete realizations of the irreducible representations, we already have a scalar product and an orthonormal basis $Y_\ell^m$ (see \nameref{sec:spherical_harmonics}). On the tensor product $V \otimes W$ of vector spaces $V$ and $W$ with orthonormal basis vectors $\vv e_i$ and $\vv f_j$, there is a unique scalar product on $V \otimes W$ such that the $\vv e_i \otimes \vv f_j$ are again an orthonormal basis (see the definition of the tensor product at the start of this section). Equivalently, it can also be defined by the property $\langle \vv v \otimes \vv w, \vv v' \otimes \vv w'\rangle = \langle \vv v , \vv v'\rangle  \cdot \langle \vv w, \vv w'\rangle$ for $\vv v, \vv v' \in V$ and $\vv w, \vv w' \in W$. If the scalar products on $V$ and $W$ are $G$--invariant, then also this scalar product on $V \otimes W$ is $G$--invariant, and we choose our isomorphism $\irreps{a}\couples{c}\irreps{b} \arrow \irreps{c}$ such that it respects these scalar products. This fixes it up to an arbitrary choice of sign (which is not important for the considerations that follow).

\phantomsection
\paragraph[Example]{Example: Coupling vectors}
\label{sec:coupling_vectors_example}
To illustrate the meaning of these operations, consider the ``coupling'' of two vectors $\vv{u},\vv{v}\in\BR^3$. Their tensor product is given by
\begin{equation*}
   \vv{u} \otimes \vv{v} =
   \begin{bmatrix} u_{x} \\ u_{y} \\ u_{z} \end{bmatrix} \otimes
   \begin{bmatrix} v_{x} \\ v_{y} \\ v_{z} \end{bmatrix} =
   \begin{bmatrix}
   u_{x}v_{x} & u_{x}v_{y} & u_{x}v_{z} \\
   u_{y}v_{x} & u_{y}v_{y} & u_{y}v_{z} \\
   u_{z}v_{x} & u_{z}v_{y} & u_{z}v_{z}
   \end{bmatrix}\,.
\end{equation*}
As stated before, this matrix can be written in a unique way as a sum of:
\vspace{-\topsep}
\begin{itemize}
    \item[$\irreps{0}$:] a multiple of the identity matrix $\mat I_3$,\vspace{0.25\baselineskip}
    \item[$\irreps{1}$:] an anti--symmetric matrix, and \vspace{0.25\baselineskip}
    \item[$\irreps{2}$:] a symmetric matrix with trace~$0$.
\end{itemize}
\vspace{-\topsep}
The first part ($\irreps{0}$ component) is the multiple of the $3 \times 3$ identity matrix $\mat I_3$ that has the same trace as the matrix $\vv{u} \otimes \vv{v}$:
\begin{equation*}
    \vv{u} \otimes^{(0)} \vv{v} 
    = \frac{\langle\vv{u},\vv{v}\rangle}{3} \cdot \mat I_3
    = \frac{
   u_x v_x + u_y v_y + u_z v_z}{3} \cdot \mat I_3\,.
\end{equation*}
To get a $1$--dimensional representation, we can use the isomorphism
\begin{equation*}
   \begin{bmatrix}
   1 & 0 & 0\\
   0 & 1 & 0\\
   0 & 0 & 1\\
   \end{bmatrix} \mapsto 1\,.
\end{equation*}
To also get a norm--preserving isomorphism, we need to multiply by the norm of the identity matrix $\sqrt{3}$, which gives
\begin{equation*}
   \vv{u} \otimes^{(0)} \vv{v} \mapsto \frac{1}{\sqrt{3}}\langle\vv{u},\vv{v}\rangle\,.
\end{equation*}

The second part ($\irreps{1}$ component) is the anti--symmetric matrix
\begin{equation*}
\begin{aligned}
   \vv{u}& \otimes^{(1)} \vv{v} = \frac12 \left(\vv{u}\vv{v}^\T - \vv{v}\vv{u}^\T\right)\\
   &= \frac12
   \begin{bmatrix}
   0 & u_x v_y-u_y v_x & u_x v_z - u_z v_x\\
   u_y v_x - u_x v_y & 0 & u_y v_z-u_z v_y\\
   u_z v_x - u_x v_z & u_z v_y - v_z u_y & 0
   \end{bmatrix}
\end{aligned}
\end{equation*}
which under the isomorphism \eqref{eq:antisymm_3x3_matrix} is mapped to the $3$--dimensional representation
\begin{equation*}
\frac{1}{2}
   \begin{bmatrix}
   u_y v_z - u_z v_y\\
   u_z v_x - u_x v_z\\
   u_x v_y - u_y v_x
   \end{bmatrix} = \frac12 \left(\vv{u}\times \vv{v}\right)\,.
\end{equation*}
This isomorphism maps e.g.\
\begin{equation*}
   \begin{bmatrix}
    0 & 1 & 0\\
   -1 & 0 & 0\\
    0 & 0 & 0
   \end{bmatrix}
   \mapsto
   \begin{bmatrix}  0\\   0\\   1
   \end{bmatrix}\,,
\end{equation*}
so to respect the norms, we have to multiply by a factor $\sqrt{2}$, 
which gives
\begin{equation*}
 \vv{u} \otimes^{(1)} \vv{v} \mapsto \frac{1}{\sqrt{2}} (\vv u \times \vv v)\,.
\end{equation*}

Finally, the part that is a traceless symmetric matrix ($\irreps{2}$ component) can be written as
\begin{equation*}
\begin{aligned}
\vv{u} \otimes^{(2)} \vv{v} = &\frac12 \left(\vv{u}\vv{v}^\T + \vv{v}\vv{u}^\T\right) - \frac{1}{6} \mathrm{Tr}\left(\vv{u}\vv{v}^\T + \vv{v}\vv{u}^\T\right)\cdot \mat I_3\\
\coloneqq &\ \mat S =
\begin{bmatrix}
S_{xx} & S_{xy} & S_{xz} \\
S_{xy} & S_{yy} & S_{yz} \\
S_{xz} & S_{yz} & S_{zz}
\end{bmatrix}\,,
\end{aligned}
\end{equation*}
where $\mathrm{Tr}$ is the trace. Using the isomorphism \eqref{eq:quadratic_forms}, the matrix $\mat S$ can be mapped to the function
\begin{equation}
\begin{aligned}
  &S_{xx} \cdot x^2 + S_{yy} \cdot y^2 + S_{zz} \cdot z^2\\
  &\quad+2 S_{xy} \cdot xy + 2 S_{xz} \cdot xz + 2 S_{yz} \cdot yz\,.
\label{eq:symmetric_matrix_as_function}
\end{aligned}
\end{equation}
Since $\mathrm{Tr}(\mat S) = S_{xx} + S_{yy} + S_{zz}=0$, this is a harmonic polynomial, and we can write it as a linear combination of our basis (see \nameref{sec:spherical_harmonics}):
\begin{equation*}
\begin{aligned}
 Y_2^{2} &= \sqrt{\frac{5}{4\pi}} \cdot \frac{\sqrt{3}}{2}\cdot\left(x^2 - y^2\right)\\
 Y_2^{-2} &= \sqrt{\frac{5}{4\pi}} \cdot\sqrt{3} \cdot x y\\
 Y_2^{1} &= \sqrt{\frac{5}{4\pi}} \cdot\sqrt{3} \cdot x z\\
 Y_2^{-1} &= \sqrt{\frac{5}{4\pi}} \cdot\sqrt{3} \cdot y z\\
 Y_2^{0} &= \sqrt{\frac{5}{4\pi}} \cdot \left(z^2-\frac{x^2}{2}-\frac{y^2}{2}\right)\,.
\end{aligned}
\end{equation*}

We now want to find the coefficients for expressing \eqref{eq:symmetric_matrix_as_function} in the orthonormal basis $Y_2^2,Y_2^{-2},\dots,Y_2^0$. For the first three terms in \eqref{eq:symmetric_matrix_as_function}, we get (using $S_{xx} + S_{yy} + S_{zz}=0$)
\begin{equation*}
\begin{aligned}
&S_{xx}\cdot x^2 + S_{yy} \cdot y^2 + S_{zz} \cdot z^2\\
&\quad= \sqrt{\frac{4\pi}{15}} \left( \left(S_{xx} - S_{yy}\right) \cdot Y_2^2 + \sqrt{3}S_{zz} \cdot Y_2^0
\right)\,,
\end{aligned}
\end{equation*}
and for the last three terms, we get
\begin{equation*}
\begin{aligned}
&2 S_{xy} \cdot xy + 2 S_{xz} \cdot xz +  2 S_{yz} \cdot yz\\
&\quad= \sqrt{\frac{4 \pi}{15}} \left(2 S_{xy} \cdot Y^{-2}_2 + 2 S_{xz} \cdot Y^{1}_2 + 2 S_{yz} \cdot Y^{-1}_2 \right)\,.
\end{aligned}
\end{equation*}
Put together, we have mapped the traceless symmetric matrix $\mat S$ to the $5$--dimensional vector
\begin{equation*}
\mat S \ \mapsto\  \sqrt{\frac{4\pi}{15}} \begin{bmatrix}
   S_{xx} - S_{yy}\\
   2 S_{xy}\\
   2 S_{xz}\\
   2 S_{yz}\\
   \sqrt{3} S_{zz}
   \end{bmatrix}
\end{equation*}
(expressed in the basis $Y_2^2,Y_2^{-2},\dots,Y_2^0$).

To correct the isomorphism used here (to make it also respect the norms), we can compute this mapping for an example of a simple traceless symmetric matrix
\begin{equation*}
\begin{bmatrix}
0 & 1 & 0 \\
1 & 0 & 0 \\
0 & 0 & 0
\end{bmatrix}\\
 \mapsto
 \sqrt{\frac{4\pi}{15}} 
 \begin{bmatrix}0\\2\\0\\0\\0\end{bmatrix}
\end{equation*}
and observe that the norm picked up a factor of $\sqrt{\frac{8\pi}{15}}$. Thus, dividing by this factor gives the norm-preserving map
\begin{equation*}
 \mat S \ \mapsto\  \frac{1}{\sqrt{2}} \begin{bmatrix}
   S_{xx} - S_{yy}\\
   2 S_{xy}\\
   2 S_{xz}\\
   2 S_{yz}\\
   \sqrt{3} S_{zz}
   \end{bmatrix}\,,
\end{equation*}
or, re-written directly in terms of the components of $\vv{u}$ and $\vv{v}$,
\begin{equation*}
\vv{u} \otimes^{(2)} \vv{v} \mapsto \frac{1}{\sqrt{2}} \begin{bmatrix}
   u_x v_x - u_y v_y\\
   u_x v_y + u_y v_x\\
   u_x v_z + u_z v_x\\
   u_y v_z + u_z v_y\\
  \frac{1}{\sqrt{3}} \left(2 u_z v_z - u_x v_x - u_y v_y\right)
   \end{bmatrix}\,.
\end{equation*}

\phantomsection
\paragraph{Clebsch--Gordan coefficients}
\label{sec:clebsch_gordan_coefficients}

The crucial observation in the above formulas is that all components of the irreps can be written as linear combinations of the components of the tensor product $\vv{u} \otimes \vv{v}$ with appropriate coefficients (and vice versa).

In general, assume we are given \mbox{$\vv{u}\in\BR^{2\ell_1+1}$} and \mbox{$\vv{v}\in\BR^{2\ell_2+1}$} and want to compute the irrep \mbox{$\vv{w}\in\BR^{2\ell_3+1}$} in the direct sum representation of the tensor product $\vv{u}\otimes\vv{v}$. As in our examples above, we fix the isomorphism such that it also respects the scalar products/norms inside the tensor product and in our concrete realization of the irreducible representations.

Using the same notation as for the spherical harmonics in \eqref{eq:spherical_harmonics}, we label the components of $\vv{u}$ and $\vv{v}$ as $u_{\ell_1}^{m_1}$ and $v_{\ell_2}^{m_2}$.\footnote{For example, consider the $3$\nobreakdash--dimensional vector \mbox{$\vv{u} = \begin{bmatrix} u_{x}\ u_{y}\ u_{z}\end{bmatrix}^\T$}. In the spherical harmonics convention, we re-label the components as $u_{x}=u_{1}^{1}$, $u_{y}=u_{1}^{-1}$, and $u_{z}=u_{1}^{0}$. This way, there is a consistent labelling scheme that works for the components of any $(2\ell+1)$\nobreakdash--dimensional irrep.} Then, the components of $\vv{w}$ are given by
\begin{equation}
w_{\ell_3}^{m_3} = \sum_{m_1=-\ell_1}^{\ell_1}  \sum_{m_2=-\ell_2}^{\ell_2} C_{\ell_1,m_1,\ell_2,m_2}^{\ell_3,m_3} u_{\ell_1}^{m_1}v_{\ell_2}^{m_2}\,,
\label{eq:cg_coupling}
\end{equation}
where $C_{\ell_1,m_1,\ell_2,m_2}^{\ell_3,m_3}$ are the so-called \emph{Clebsch--Gordan coefficients} (CGCs).

For example, the CGCs necessary for computing $\irreps{1}\couples{0}\irreps{1}$ with \eqref{eq:cg_coupling} are
\begin{equation*}
C_{1,-1,1,-1}^{0,0} = C_{1,0,1,0}^{0,0} = C_{1,1,1,1}^{0,0} = \frac{1}{\sqrt{3}}
\end{equation*}
and 
\begin{equation*}
\begin{aligned}
&C_{1,-1,1,0}^{0,0} = C_{1,-1,1,1}^{0,0} = C_{1,0,1,-1}^{0,0}\\ &\quad = C_{1,0,1,1}^{0,0} = C_{1,1,1,-1}^{0,0} = C_{1,1,1,0}^{0,0} = 0
\end{aligned}
\end{equation*}
(compare to the norm-preserving isomorphism given  for the $\irreps{0}$ component in the \hyperref[sec:coupling_vectors_example]{example for coupling vectors} above).

As mentioned before, our construction using norm--preserving isomorphisms only determines the CGCs up to an arbitrary choice of sign per triple $(\ell_1, \ell_2, \ell_3)$. The choice realized in \texttt{E3x} has the property that $C_{\ell_1,0,\ell_2,0}^{\ell_3,0}\geq0$.

\phantomsection
\subsection[Representations of \texorpdfstring{$\Othree$}{O(3)}]{Representations of $\Othree$}
\label{sec:irreps_of_o3}
So far, we have mostly focused on representations for $\SOthree$, we now want to infer the corresponding results for $\Othree$. At this point, we clarify that the equivariant features used in \texttt{E3x} are built from irreps of $\Othree$. In the remainder of this manuscript, when the generic term ``irreps'' is used without specifying a particular group, we implicitly mean irreps of $\Othree$.

The group $\Othree$ consists of rotations $g\in \SOthree$ and rotoreflections (see also Table~\ref{tab:groups_relevant_in_3d}), which can be written as $-e \odot g$ for a $g\in \SOthree$ ($e$ is the identity element). Since $-e \odot g = g \odot -e = -g$, a representation $(\rho, V)$ of $\Othree$ is given by the values $\rho(g)$ for $g\in \SOthree$ (i.e.\ a representation of $\SOthree$ on $V$) and one additional matrix $\rho(-e)$ with the property that $\rho(-e)^2 = \rho\left((-e)^2\right) = \rho(e)$ is the identity on $V$ and commutes\footnote{The phrase ``$g,h$ commute'' means $g\odot h = h \odot g$.} with all $\rho(g)$ for $g\in \SOthree$. So for every representation $(\rho, V)$ of $\SOthree$, there are two representations $(\rho_{\pm}, V)$ of $\Othree$ given by $\rho(-e)=\pm \mat I_n$ ($\mat I_n$ denotes the $n\times n$ identity matrix). In particular we get irreducible representations $\irrep{0}{\pm}, \irrep{1}{\pm},...$ of $\Othree$. These are in fact \emph{all} irreducible representations of $\Othree$.\footnote{For reducible $\SOthree$--representations, there are more ways to extend them to $\Othree$--representations: We can choose a sign for each irreducible component.} We refer to $\Othree$--representations with $\rho(-e)=\mat I_n$ as \emph{even} (or ``parity~$+1$'') and with $\rho(-e)=-\mat I_n$ as \emph{odd} (or ``parity~$-1$'').

For example, the trivial $1$\nobreakdash--dimensional representation is $\irrep{0}{+}$, and the canonical $3$\nobreakdash--dimensional representation is $\irrep{1}{-}$. These irreps correspond to scalars and vectors, whereas the irreps $\irrep{0}{-}$ and $\irrep{1}{+}$ correspond to ``pseudoscalars'' and ``pseudovectors'',\footnote{For example, the cross product $\vv{w} = \vv{u} \times \vv{v}$ of two vectors $\vv{u}$ and $\vv{v}$ is a pseudovector. Under a reflection (refl.) of the coordinate system, (proper) vectors change direction, i.e.\ $\vv{u} \xrightarrow{\text{refl.}} -\vv{u}$ and $\vv{v} \xrightarrow{\text{refl.}} -\vv{v}$, but pseudovectors do not, i.e.\ $\vv{w} \xrightarrow{\text{refl.}} \vv{w}$. Similarly, the dot product $\vv{u} \cdot \vv{v}$ of two vectors is a scalar (which does not change sign under reflection), whereas the dot product of a vector and a pseudovector is a pseudoscalar (which does change sign under reflection).} respectively. In general, we refer to irreps with even/odd degree and even/odd parity as (proper) tensors, whereas irreps with even/odd degree and odd/even parity are referred to as ``pseudotensors''.

For tensor products of even/odd representations, the parities multiply, so the general rule for expressing the tensor product $\irrep{a}{\alpha} \otimes \irrep{b}{\beta}$ of two irreps $\irrep{a}{\alpha}$ and $\irrep{b}{\beta}$ of $\Othree$ as a direct sum is
\begin{equation*}
 \irrep{a}{\alpha} \otimes \irrep{b}{\beta} \simeq
   \begin{cases}
    \irrep{\left(\lvert a-b \rvert\right)}{+}  \oplus
   \dots \oplus
    \irrep{\left(a+b \right)}{+} & \alpha = \beta \\
    \irrep{\left(\lvert a-b \rvert\right)}{-}  \oplus
   \dots \oplus
    \irrep{\left(a+b \right)}{-} & \alpha \neq \beta\,. \\
   \end{cases}
\end{equation*}
Following the short-hand notation we introduced earlier, we also write the $\irrep{c}{\gamma}$--component of $\irrep{a}{\alpha} \otimes \irrep{b}{\beta}$ as $\irrep{a}{\alpha} \couple{c}{\gamma} \irrep{b}{\beta}$, which is computed with \eqref{eq:cg_coupling} in the same way as couplings of $\SOthree$--irreps. The CGCs for computing the coupling of $\Othree$--irreps $\irrep{a}{\alpha} \couple{c}{\gamma} \irrep{b}{\beta}$ are identical to the ones for computing the coupling of $\SOthree$--irreps $\irreps{a} \couples{c} \irreps{b}$ when $\gamma = \alpha \cdot \beta$ (and zero otherwise).

\phantomsection
\section{How \texttt{E3x} works}
\label{sec:how_e3x_works}

The core design principle of \texttt{E3x} is that implementing an equivariant deep learning architecture should feel familiar to anyone who has experience with constructing ordinary neural networks. Loosely speaking, features in \texttt{E3x} are ``augmented with directional information'', and components like activation functions or linear layers are generalized to work with such features while preserving their equivariance. Importantly, ``ordinary features'' can be recovered as a special case, in which case the ``ordinary behaviour'' of common neural network components is recovered as well. By keeping standard neural network building blocks essentially unchanged, it is straightforward to define equivariant models with almost no necessary code changes compared to ordinary models, see Listing~\ref{code:flax_vs_e3x} for an example.

\begin{lstlisting}[float=*, floatplacement=tbp, language=Python, caption=Simple multi-layer perceptron (MLP) written with Flax\cite{flax2020github} compared to an equivariant MLP written with \texttt{E3x}., label=code:flax_vs_e3x]
# 2-layer ordinary MLP with flax.
from flax import linen as nn
...
h = nn.Dense(features=num_hidden)(x)  # x stores (invariant) ordinary features
y = nn.Dense(features=num_output)(nn.relu(x))

# 2-layer equivariant MLP with e3x.
import e3x
...
h = e3x.nn.Dense(features=num_hidden)(x)  # x stores equivariant irrep features
y = e3x.nn.Dense(features=num_output)(e3x.nn.relu(x))
\end{lstlisting}

In the following, we give an overview over the basic components of \texttt{E3x}, including equivariant features and how familiar neural network building blocks, such as dense layers and activation functions, are generalized to work with them. We also describe tensor layers, which couple equivariant features in a way that has no analogue in ordinary neural networks (see also \nameref{sec:coupling_irreps}). Finally, we briefly mention other utilities in \texttt{E3x} that can be useful when working with three--dimensional data.

\begin{figure}[tbp]
    \centering
    \includegraphics{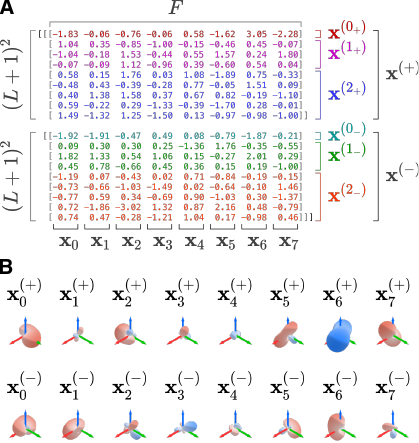}
    \caption{(\textbf{A}) Color-coded visualization of the memory layout of (randomly drawn) features \mbox{$\vv{x}\in\BR^{2\times(L+1)^2\times F}$} with $L=2$ and $F=8$. (\textbf{B}) Visualization of features as three-dimensional shapes (negative components are drawn in red and positive components in blue).}
    \label{fig:feature_memory_layout}
\end{figure}

\phantomsection
\paragraph{Irrep features}
\label{sec:irrep_features}
Ordinary neural networks typically operate on features $\vv{x}\in\BR^F$, where $F$ is the dimension of the feature space. We can either think of $\vv{x}$ as an $F$\nobreakdash--dimensional feature vector, or equivalently, as a collection of $F$ scalar ($1$\nobreakdash--dimensional) features. In \texttt{E3x}, a single feature is not necessarily just a scalar anymore, but instead consists of irreps of $\Othree$ (see \nameref{sec:irreps_of_o3}) for all degrees starting from $\ell=0$ up to some maximum degree $\ell=L$ (we call these \emph{irrep features}). Since irreps of degree $\ell$ have $2\ell+1$ components and there is an irrep with both even and an odd parity for each $\ell$, the features in E3x are $\vv{x}\in\BR^{2\times(L+1)^2\times F}$ in the most general case (see also Fig.~\ref{fig:spherical_harmonics}, where all $(L+1)^2$ spherical harmonics up to a maximum degree $L$ are visualized in a square arrangement). We use the short-hand notation $\vv{x}_i$ to refer to all irreps of the $i$-th feature, $\vv{x}^{(\ell_\pm)}$ to refer to all
irreps of degree $\ell$ with parity $p = \pm 1$, and
$\vv{x}^{(\pm)}$ to refer to all irreps of all degrees with parity
$p = \pm 1$. In our implementation, $\vv{x}$ is stored in a contiguous array \texttt{x} of shape \texttt{(2, (L+1)**2, F)} (see Fig.~\ref{fig:feature_memory_layout}A).\footnote{This has the advantage that the coupling of irrep features via CGCs (see \eqref{eq:cg_coupling}) can be efficiently implemented on accelerators such as GPUs and TPUs with \texttt{einsum} operations.} For reference, an overview of how to ``translate'' between the short-hand notation introduced above and the corresponding array slicing operation in code (in the style of \texttt{numpy}\cite{harris2020array}) is given in Table~\ref{tab:slicing_notation}. 

\begin{table*}[tbp]
\centering
\begin{tabular}{l l l}
\toprule
\textbf{notation} & \texttt{numpy}\textbf{-style array slicing} & \textbf{shape} \\
\midrule
   $\vv{x}$ & \texttt{x} & \texttt{(2, (L+1)**2, F)}\\
   $\vv{x}_i$ & \texttt{x[:, :, i:i+1]} & \texttt{(2, (L+1)**2, 1)} \\ 
   $\vv{x}^{(+)}$ & \texttt{x[0:1, :, :]} & \texttt{(1, (L+1)**2, F)}\\
   $\vv{x}^{(-)}$ &  \texttt{x[1:2, :, :]} & \texttt{(1, (L+1)**2, F)}\\
   $\vv{x}^{(\ell_+)}$  & \texttt{x[0:1, l**2:(l+1)**2, :]} & \texttt{(1, 2*l+1, F)}\\
   $\vv{x}^{(\ell_-)}$ & \texttt{x[1:2, l**2:(l+1)**2, :]} & \texttt{(1, 2*l+1, F)}\\
\bottomrule
\end{tabular}
\caption{Correspondence between mathematical short-hand notation for specific components of $\vv{x}$ and \texttt{numpy}-style array slicing.}
\label{tab:slicing_notation}
\end{table*}

Another way to think about irrep features is to imagine them as representing abstract three--dimensional shapes with positive and negative regions (see Fig.~\ref{fig:feature_memory_layout}B). Under rotations of the coordinate system, these shapes rotate accordingly; under reflections, all regions of features with negative parity, i.e.\ $\vv{x}^{(-)}$, change sign (features with positive parity stay unchanged under reflections). Crucially, as long as the maximum degree $L$ is chosen sufficiently large, all possible ``behaviors under rotations and reflections'' can be expressed by such irrep features (see \nameref{sec:irreps}), i.e.\ all equivariant quantities can be ``assembled'' from irrep features. 

\begin{figure}[ht]
    \centering
    \includegraphics{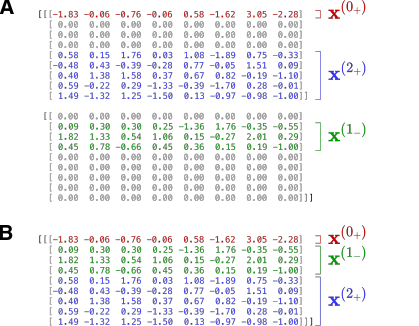}
    \caption{For better memory efficiency, \texttt{E3x} supports representing features where all pseudotensor components are zero (\textbf{A}) without the need for explicitly storing the zero components (\textbf{B}).}
    \label{fig:no_pseudotensor_features}
\end{figure}

Since it is often useful to work with irrep features where all pseudotensors are zero and only ``proper tensor'' components are non-zero, \texttt{E3x} also supports representing such features without the need to explicity store the zero components (see Fig.~\ref{fig:no_pseudotensor_features} for an illustration). All operations implemented in \texttt{E3x} automatically detect which kind of features are used (from their shape) and computations that involve features without any pseudotensor components are (apart from being implemented more efficiently) equivalent to using features where all pseudotensor components are set to zero. Thus, for brevity, we only mention the general case of irrep features that contain both tensor and pseudotensor components in the descriptions that follow. 

It is worth pointing out that when pseudotensors are omitted and the maximum degree is $L=0$ (i.e.\ there are only scalar feature components), irrep features become equivalent to the features used in ordinary neural networks. Thus, \texttt{E3x} implements a generalization that allows to define equivariant and ordinary models with the same code (the maximum degree $L$ of irrep features can be a hyperparameter).

\phantomsection
\paragraph{Activation functions}
\label{sec:activation_functions}

An essential component of neural networks are non-linear activation functions, which are typically applied element-wise to (scalar) features. For the irrep features used in \texttt{E3x}, an element-wise application of a non-linear function would not preserve equivariance, because the absolute value and sign of the individual feature entries depends on the orientation of the chosen coordinate system. However, multiplying all elements with a common (scalar) factor \emph{does} preserve equivariance. Thus, given a non-linear function $\sigma(x)$, we can define an equivariance-preserving application of $\sigma$ to irrep features $\vv{x}$ as
\begin{equation}
\sigma(\vv{x})\coloneqq g_\sigma\left(\vv{x}^{(0_+)}\right) \circ \vv{x}\,.
\label{eq:gated_linear_activation}
\end{equation}
Here, `$\circ$' denotes element-wise multiplication (with implied broadcasting\footnote{Broadcasting means that dimensions that are size 1 in one of the factors are implicitly repeated to match the size of the corresponding dimension in the other factor.}), and $g_\sigma$ (which is applied element-wise to only the scalar feature components $\vv{x}^{(0_+)}$) satisfies
\begin{equation}
\sigma(x)=g_\sigma(x)\cdot x\,.
\label{eq:gating_function}
\end{equation}
Evidently, when pseudotensors are omitted and $\ell=0$ (i.e,\ there are only scalar feature components), \eqref{eq:gated_linear_activation} is equivalent to the familiar element-wise application of $\sigma$ for ordinary features. Many common activation functions can be transformed to the form given in \eqref{eq:gating_function}, for example
\begin{equation*}
\mathrm{relu}(x) = \max(0, x)
\Rightarrow g_{\mathrm{relu}}(x)\ =\ \max(0, \mathrm{sgn}\ x)
\end{equation*}
or
\begin{equation*}
\mathrm{swish}(x) = \frac{x}{1+e^{-x}}
\ \Rightarrow\ g_{\mathrm{swish}}(x) = \frac{1}{1+e^{-x}}
\end{equation*}
(\texttt{E3x} already contains implementations of~\eqref{eq:gated_linear_activation} for most popular activation functions). Implementing new activation functions is typically straightforward and requires only a few lines of code, see Listing~\ref{code:relu_implementation} for an example. 

\begin{lstlisting}[float=*, floatplacement=tbp, language=Python, caption=Implementation of an equivariant generalization of the popular $\mathrm{relu}(x)$ activation function following the general form given in \eqref{eq:gated_linear_activation}., label=code:relu_implementation]
import jax.numpy as jnp

def relu(x):
  return jnp.maximum(jnp.sign(x[..., 0:1, 0:1, :]), 0.0) * x
\end{lstlisting}

\phantomsection
\paragraph{Dense layers}
\label{sec:dense_layers}
Dense (or fully-connected) layers are perhaps the most basic building block of neural networks. Due to their linearity, it is straightforward to generalize them to support irrep features: As mentioned before, multiplying all elements of an irrep with a common (scalar) factor preserves equivariance. Similarly, element-wise addition of irreps of the same kind (sharing the same parity and degree) is also an equivariant operation. It follows that (scalar) bias terms may only be added to scalar feature components. Thus, for $\vv{y} = \mathrm{dense}(\vv{x})$, where $\vv{x}\in\BR^{2 \times (L+1)^2 \times F_{\mathrm{in}}}$ and $\vv{y}\in\BR^{2 \times (L+1)^2 \times F_{\mathrm{out}}}$ are the input and output features, the components of $\vv{y}$ with degree $\ell$ and parity $p$ are given by
\begin{equation*}
\vv{y}^{(\ell_p)} = \begin{cases}
\vv{x}^{(0_+)}\mat{W}_{(0_+)} + \vv{b} & \ell=0,\ p=+1\\
\vv{x}^{(\ell_p)}\mat{W}_{(\ell_p)} & \mathrm{otherwise}\,.
\end{cases}
\end{equation*}
Here, $\mat{W}\in \BR^{F_{\mathrm{in}}\times F_{\mathrm{out}}}$ are weights, $\vv{b}\in \BR^{F_{\mathrm{out}}}$ biases, and all operations are implicitly broadcasted across dimensions.\footnote{The notation $\vv{x}^{(\ell_p)}\mat{W}$ means that each $F_{\mathrm{in}}$\nobreakdash--dimensional ``slice'' of $\vv{x}^{(\ell_p)}\in\BR^{2 \times (2\ell+1) \times F_{\mathrm{in}}}$ is multiplied with $\mat{W} \in \BR^{F_{\mathrm{in}}\times F_{\mathrm{out}}}$ according to the ordinary matrix-vector multiplication rules. Similarly, $\vv{b} \in \BR^{F_{\mathrm{out}}}$ is added element-wise to each slice of $\vv{x}^{(0_+)}\mat{W}$.} While it would be possible to use the same weights for feature components of all degrees and parities, in \texttt{E3x}, every irrep of the output features has a separate weight matrix $\mat{W}_{(\ell_p)}$, i.e.\ there are \mbox{$2(L+1)F_{\mathrm{in}}F_{\mathrm{out}}$} weight parameters in total.\footnote{Unless the input features $\vv{x}$ do not contain any pseudotensor components (see \nameref{sec:irrep_features}), in which case the corresponding parameters are omitted and there are only \mbox{$(L+1)F_{\mathrm{in}}F_{\mathrm{out}}$} weight parameters.} Separate weight matrices make the equivariant generalization of dense layers more expressive, because they allow to change the proportion of different irreps within each feature $\vv{y}_i$ independently.

\phantomsection
\paragraph{Tensor layers}
\label{sec:tensor_layers}
As described in \nameref{sec:coupling_irreps}, it is possible to ``couple'' different irreps via tensor products to produce new irreps. This operation is at the heart of building equivariant architectures and has no proper analogue in ordinary neural networks (although for purely scalar features, it is equivalent to element-wise multiplication). The result of a learnable tensor product coupling $\vv z = \mathrm{tensor}(\vv x, \vv y)$ of irrep features $\vv x \in \BR^{2\times(L_x+1)^2\times F}$ and $\vv y \in \BR^{2\times(L_y+1)^2\times F}$ is computed as
\begin{equation*}
\vv{z}^{(c_\gamma)} = \sum_{(a_\alpha,b_\beta)}
      \vv{w}_{(a_\alpha,b_\beta,c_\gamma)} \circ \left(
      \vv{x}^{(a_\alpha)} \otimes^{(c_\gamma)}\vv{y}^{(b_\beta)}
      \right)\,.
\end{equation*}
Here, $\vv w \in \BR^{F}$ are weight coefficients, `$\circ$' denotes element-wise multiplication (with implied broadcasting), and the sum runs over all combinations of degrees and parities. The coupling operator `$\otimes^{(c_\gamma)}$' is implicitly broadcasted across the feature dimension, i.e.\ irreps within each feature channel are coupled independently of those in other feature channels. Each valid ``coupling path'', denoted as a triplet of degree and parity combinations $(a_\alpha,b_\beta,c_\gamma)$ has separate trainable weights. The input features $\vv x$ and $\vv y$ may have different maximum degrees $L_x$ and $L_y$, and the maximum degree $L_z$ of the output features $\vv z \in \BR^{2\times(L_z+1)^2\times F}$ can be freely chosen between $L_z = 0$ and $L_z = L_x + L_y$ (see the rule for coupling irreps in \nameref{sec:irreps_of_o3}, all irreps with degrees $\ell > L_z$ are omitted).

\phantomsection
\paragraph{Tensor dense layers}
\label{sec:tensor_dense_layers}
As described above, dense layers allow to recombine information in different feature channels, but do not mix information from different parity and degree channels. In contrast, tensor layers combine information from different parity and degree channels without mixing across the feature dimension. Thus, both types of layers can be combined to a flexible feature transformation that allows mixing across all channels:
\begin{equation*}
\begin{aligned}
\vv a &= \mathrm{dense}_1(\vv x)\\
\vv b &= \mathrm{dense}_2(\vv x)\\
\vv y &= \mathrm{tensor}(\vv a, \vv b)\,.
\end{aligned}
\end{equation*}
Here, the input features $\vv x \in \BR^{2\times (L_{\mathrm{in}}+1)^2 \times F_{\mathrm{in}}}$ are first projected to the intermediate features $\vv a, \vv b \in \BR^{2\times (L_{\mathrm{in}}+1)^2 \times F_{\mathrm{out}}}$ with separate dense layers, which are then coupled with a tensor layer to produce the output features $\vv y \in \BR^{2\times (L_{\mathrm{out}}+1)^2 \times F_{\mathrm{out}}}$.  Because they allow information flow between all feature channels, such \emph{tensor dense layers} are a powerful building block for equivariant model architectures. For example, by stacking multiple tensor dense layers on top of each other, the scalar feature components (which are invariant under rotations and reflections of the coordinate system) are successively ``enriched'' with higher-order geometric information from feature components with $\ell > 0$. Contrary to stacks of dense layers, which are mathematically equivalent to a single dense layer when not interleaved with non-linear activation functions, tensor dense layers can be stacked directly, because the tensor operation already acts as a non-linearity.


\phantomsection
\paragraph{Other layers and utilities}
\label{sec:other_layers_and_utilities}
\texttt{E3x} also includes ready-to-use equivariant implementations of more sophisticated neural network building blocks, e.g.\ message-passing\cite{gilmer2017neural} or attention.\cite{vaswani2017attention} Further, there is code for evaluating spherical harmonics, and, more generally ``featurizing'' vectors $\vv{r} \in \BR^3$ using functions of the form given in \eqref{eq:radial_so3}. For testing whether specific transformations are equivariant, \texttt{E3x} includes functionality for generating random rotation matrices and converting rotation matrices to the corresponding Wigner-D matrices (used to rotate irrep features directly). Since point clouds are a natural representation for many different types of three--dimensional data, \texttt{E3x} also contains utilities for working with point clouds. In particular, this includes support for both ``sparse'' and ``dense'' neighbor/index lists (following the nomenclature used in JAX MD\cite{schoenholz2020jax}) and ``indexed operations'', for example to compute a sum over specific points according to a given index list. For a full overview of features implemented in \texttt{E3x}, we refer the reader to the online documentation.

\section*{How to get started}
E3x is available on PyPI and can be installed by simply running:
\begin{verbatim}
python -m pip install --upgrade e3x
\end{verbatim}
Usage examples and detailed documentation can be found on \url{https://e3x.readthedocs.io}. The code for \texttt{E3x} is available from \url{https://github.com/google-research/e3x}.

\section*{Acknowledgements}
We thank J.\ T.\ Frank, N.\ W.\ A.\ Gebauer and S.\ Ganscha for helpful comments on the manuscript.

\bibliography{main}

\end{document}